\documentclass[10pt,twocolumn,letterpaper]{article}

\usepackage[pagenumbers]{cvpr} 

\usepackage{graphicx}
\usepackage{amsmath}
\usepackage{amssymb}
\usepackage{booktabs}
\usepackage{bm}

\usepackage{multirow}

\usepackage{times}
\usepackage{epsfig}
\usepackage{mathtools}
\usepackage{mathabx}
\usepackage[dvipsnames]{xcolor}
\usepackage{float}
\usepackage[export]{adjustbox}
\usepackage{caption, subcaption}
\usepackage[percent]{overpic}

\usepackage[pagebackref,breaklinks,colorlinks]{hyperref}

\usepackage[accsupp]{axessibility} 

\usepackage[capitalize]{cleveref}
\crefname{section}{Sec.}{Secs.}
\Crefname{section}{Section}{Sections}
\Crefname{table}{Table}{Tables}
\crefname{table}{Tab.}{Tabs.}

\def\confName{CVPR}
\def\confYear{2022}

\begin{document}

\title{CAT-Det: Contrastively Augmented Transformer for \\ Multi-modal 3D Object Detection}

\author{Yanan Zhang\textsuperscript{\rm 1,2}, Jiaxin Chen\textsuperscript{\rm 2}, Di Huang\textsuperscript{\rm 1,2}\thanks{indicates the corresponding author.}\\
\textsuperscript{\rm 1}State Key Laboratory of Software Development Environment, Beihang University, Beijing, China\\
\textsuperscript{\rm 2}School of Computer Science and Engineering, Beihang University, Beijing, China\\
{\tt\small \{zhangyanan, jiaxinchen, dhuang\}@buaa.edu.cn}
}
\maketitle

\begin{abstract}

In autonomous driving, LiDAR point-clouds and RGB images are two major data modalities with complementary cues for 3D object detection. However, it is quite difficult to sufficiently use them, due to large inter-modal discrepancies. To address this issue, we propose a novel framework, namely Contrastively Augmented Transformer for multi-modal 3D object Detection (CAT-Det). Specifically, CAT-Det adopts a two-stream structure consisting of a Pointformer (PT) branch, an Imageformer (IT) branch along with a Cross-Modal Transformer (CMT) module. PT, IT and CMT jointly encode intra-modal and inter-modal long-range contexts for representing an object, thus fully exploring multi-modal information for detection. Furthermore, we propose an effective One-way Multi-modal Data Augmentation (OMDA) approach via hierarchical contrastive learning at both the point and object levels, significantly improving the accuracy only by augmenting point-clouds, which is free from complex generation of paired samples of the two modalities. Extensive experiments on the KITTI benchmark show that CAT-Det achieves a new state-of-the-art, highlighting its effectiveness.

\end{abstract}

\section{Introduction}
\label{sec:intro}

3D object detection is a fundamental step in autonomous driving perception systems. It mainly operates 3D point-clouds acquired by LiDAR sensors and provides important spatial clues including location, direction, and object size. Despite true and accurate geometry information recorded, the distribution of point-clouds is disordered, irregular, and sparse, making 3D object detection a challenging task.

\begin{figure}[t]
  \centering
   \includegraphics[width=0.9\linewidth]{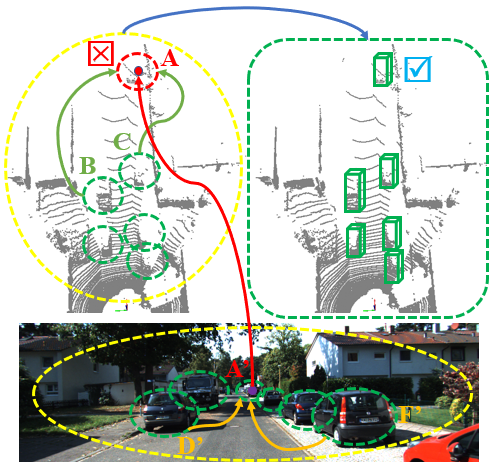} 
   \caption{Illustration of the fusion process in CAT-Det. (A): A failure case due to few points at a long distance in the point-cloud modality. (A'): Its corresponding case in the image modality. Although the feature of (A) is enhanced by those of (B) and (C) in the PT branch, this is often insufficient. With the CMT module, the feature of (A) is further enhanced by that of (A') which also integrates the contributions from (D') and (F') in the IT branch, and accurate detection is finally achieved.}
   \label{fig:1}
   \vspace{-0.4cm}
\end{figure}

The past few years have witnessed the fast development in 3D object detection. A large number of methods are introduced in the literature, and according to the input form in detection feature learning, the methods are roughly categorized into grid-based and point-based. The former initially converts point-clouds into regular grids by projecting them to images of specific views~\cite{engelcke2017vote3deep, liang2019multi} or subdividing them to voxels in the space ~\cite{wang2015voting, zhou2018voxelnet, yan2018second, lang2019pointpillars} and further conducts 2D or 3D Convolutional Neural Networks (CNN) to encode geometric cues. The latter directly takes raw points and applies point-cloud deep learning networks \emph{e.g.} PointNet~\cite{qi2017pointnet}/PointNet++~\cite{qi2017pointnet++} or graph neural networks \emph{e.g.} DGCNN~\cite{wang2019dynamic} to capture shape structures~\cite{shi2019pointrcnn, yang20203dssd, shi2020point, zhang2020pc}. More recently, several attempts~\cite{yang2019std, shi2020pv, he2020structure} deliver stronger models by integrating point-based and grid-based networks as hybrid representations, reporting better results.

To boost the performance of 3D object detection, another strategy targets on multi-modal solutions, which makes use of 3D point-clouds along with 2D images. Although images have not yet proved so competent as an independent modality for this issue evidenced by inferior baselines~\cite{chen2016monocular, chabot2017deep, mousavian20173d, li2019stereo}, the combination of geometric and textural clues conveyed in point-clouds and images does lead to accuracy gains for their natural complementarity~\cite{chen2017multi, ku2018joint, qi2018frustum, vora2020pointpainting, yoo20203d}. F-PointNet~\cite{qi2018frustum} and F-ConvNet~\cite{wang2019frustum} perform the fusion in series, where 3D frustum proposals are firstly cropped based on prepared 2D regions through a standard 2D CNN detector and each point within the proposal is then segmented and screened using a PointNet-like block for regression. By contrast, more studies fulfill this task in parallel. For example, \cite{vora2020pointpainting, xie2020pi} conduct data-level fusion by enhancing 3D coordinates with point-wise 2D segmentation features; \cite{chen2017multi, ku2018joint, liang2018deep, liang2019multi, huang2020epnet} achieve feature-level fusion of 2D and 3D representations from individual networks by simple concatenation or specific modules; and \cite{pang2020clocs} implements box-level fusion which merges the individual candidate sets of a couple of 2D and 3D detectors in a learning manner. Different from the LiDAR only methods that continuously update for more sophisticatedly designed models and more suitable training schemes in the single point-cloud modality, the multi-modal alternatives endeavour to leverage more diverse information and suggest great potential. However, as the KITTI~\cite{geiger2012we} leaderboard displays, there still exists a certain gap between the multi-modal methods and the top LiDAR only ones~\cite{zheng2021se}.

Such a gap is due to three aspects. (1) In multi-modal 3D object detection, PointNet++~\cite{qi2017pointnet++}/3D sparse convolutions~\cite{yan2018second} and 2D CNNs are principal building blocks to extract point-cloud and image features respectively. Limited by their local receptive fields, contexts cannot be comprehensively acquired from both the modalities, triggering information loss. (2) The widely adopted fusion schemes, particularly the ones at the feature-level, such as direct concatenation~\cite{chen2017multi, ku2018joint}, additional convolution~\cite{liang2018deep, liang2019multi}, and simple attention~\cite{yoo20203d, huang2020epnet}, assign no weights or coarse weights learned within limited receptive fields to different features, where crucial clues are not well highlighted. (3) Ground-truth data augmentation~\cite{yan2018second} is a common practice to facilitate LiDAR only methods; unfortunately, it is not so straightforward to apply this mechanism to multi-modal methods as augmentation in the single modality tends to cause semantic misalignment. \cite{wang2021pointaugmenting} indeed presents a cross-modal augmentation technique for paired data, but the procedure on images is cumbersome and easy to incur noise.

To address the issues mentioned above, this paper proposes a novel framework for multi-modal 3D object detection, namely Contrastively Augmented Transformer Detector (CAT-Det). It adopts a two-stream structure, consisting of a Pointformer (PT) branch, an Imageformer (IT) branch together with a Cross-Modal Transformer (CMT) module. Unlike PointNet++ and CNNs, both the PT and IT branches possess large receptive fields, which are able to respectively capture rich global context information in point-clouds and images to strengthen features of hard samples. Subsequently, the CMT module conducts cross-modal feature interaction and multi-modal feature combination, where essential cues extracted in the two modalities are sufficiently emphasized with holistically learned fine-grained weights. The integration of PT, IT, and CMT fully encodes intra-modal and inter-modal long-range dependencies as a powerful representation, thus benefiting detection performance. In addition, we propose a one-way multi-modal data augmentation (OMDA) approach through hierarchical contrastive learning, which accomplishes effective augmentation by solely performing on the point-cloud modality.

In summary, the major contributions of this paper are:

(1) We propose a novel CAT-Det framework for multi-modal 3D object detection, with a pointformer branch, an imageformer branch and a cross-modal transformer module. To the best of our knowledge, it is the first attempt that applies the transformer structure to the given task. (2) We propose a one-way data augmentation approach for multi-modal 3D object detection via hierarchical contrastive learning, significantly improving the accuracy only by augmenting point-clouds, thus free from complex generation of paired samples of the two modalities. (3) We achieve a newly state-of-the-art mAP of all the three classes on the KITTI test set in comparison to the published counterparts, and demonstrate its advantage in detecting hard objects.

\section{Related Work}

\begin{figure*}[t]
\centering
\includegraphics[width=0.9\textwidth]{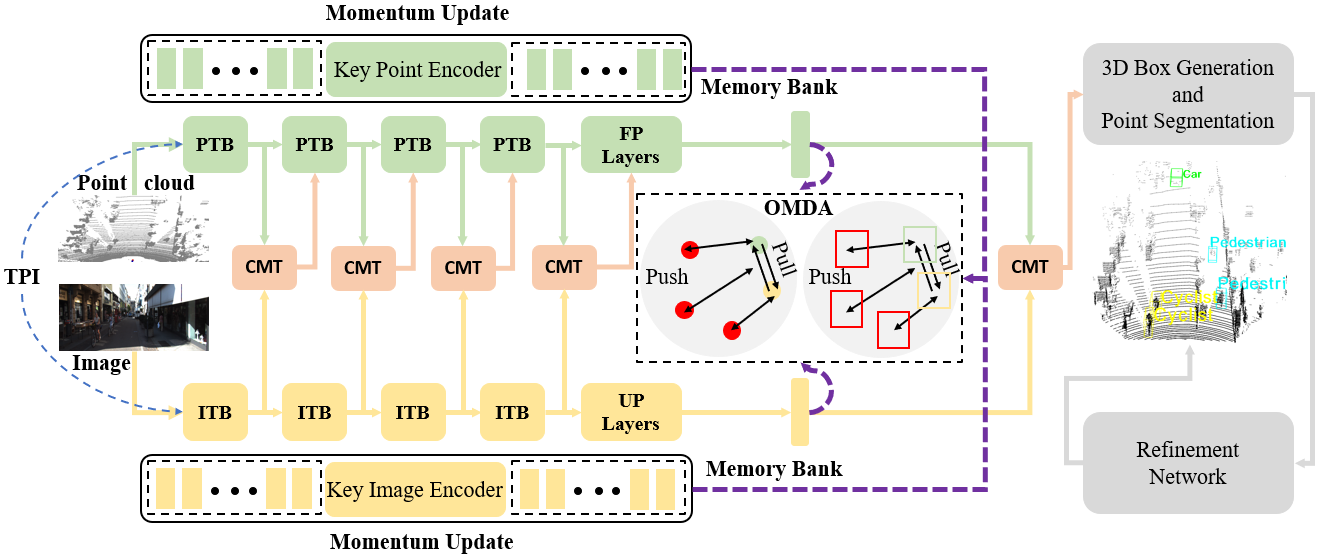} 
\caption{Framework overview. The whole framework consists of three main modules: (1) Two-stream Pointformer and Imageformer (TPI), (2) Cross-Modal Transformer (CMT), and (3) One-way Multi-modal Data Augmentation (OMDA). TPI builds intra-modal long-range context feature representation from the two modalities, and CMT performs inter-modal feature interaction and aggregation at multiple levels. In addition, OMDA conducts hierarchical contrastive learning to achieve concise yet effective data augmentation.}
\label{fig:2}
\vspace{-0.4cm}
\end{figure*}

\textbf{Image based 3D Object Detector.} Some approaches~\cite{chabot2017deep, chen20153d, chen2016monocular} perform 2D/3D matching via exhaustively sampling and scoring 3D proposals as representative templates. Numerous methods~\cite{mousavian20173d, li2019stereo, li2019gs3d, chen2020dsgn} directly start with accurate 2D bounding boxes to roughly estimate 3D pose from geometric properties obtained by empirical observation. Another way is to first conduct depth estimation and then resort to existing point-cloud based methods~\cite{bao2019monofenet, wang2019pseudo, you2019pseudo}. Although 2D object detection has made remarkable advancements, images are not regarded as a good individual modality to predict 3D objects. For the absence of depth information, monocular image based methods suffer from low precision. Stereo image based methods are able to recover depth information, but it is usually coarse with additional noise.

\textbf{Point-cloud based 3D Object Detector.} Some methods convert point-clouds to regular grids by projecting to planes~\cite{engelcke2017vote3deep,liang2019multi} or subdividing to voxels~\cite{wang2015voting, zhou2018voxelnet, yan2018second, lang2019pointpillars} so that they can be processed by 2D or 3D CNNs for feature learning. More methods take raw unordered and irregular data as input and apply point-cloud deep learning networks, such as PointNet~\cite{qi2017pointnet} and PonintNet++~\cite{qi2017pointnet++}, to encode structure features~\cite{shi2019pointrcnn, yang20203dssd} and a few methods~\cite{shi2020point,zhang2020pc} attempt graph neural networks in this step. Recent methods~\cite{yang2019std, shi2020pv, he2020structure} also utilize both point-based and voxel-based networks to extract features from different representations of point-clouds. More recently, a series of methods~\cite{liu2021group, guan2021m3detr, pan20213d, mao2021voxel, sheng2021improving, misra2021end} have emerged based on point transformer for the property to capture global contexts.

\textbf{Multi-modal 3D Object Detector.} MV3D~\cite{chen2017multi} and AVOD~\cite{ku2018joint} take LiDAR projections and RGB images as inputs and fuse region-based features to make prediction. F-PointNet~\cite{qi2018frustum} and F-ConvNet~\cite{wang2019frustum} first leverage a 2D CNN object detector to extract 2D regions from images, then transform 2D regional coordinates to the 3D space to crop frustum proposals, and finally localize interest points within the frustum by a PointNet-like block for regression. PointPainting~\cite{vora2020pointpainting} and PI-RCNN~\cite{xie2020pi} turn to semantics network for per pixel classification, and the relevant segmentation score, which serves as compact features of the image, is appended to the LiDAR points via projecting them into the segmentation mask. CLOCs~\cite{pang2020clocs} directly uses pre-trained 2D and 3D detectors by late fusion, making the proposals in different modalities connected without integrating the features. Recent studies~\cite{liang2018deep, liang2019multi, yoo20203d, huang2020epnet} combine the modalities in the feature space to obtain a multi-modal representation before feeding them into a supervised learner. Despite many efforts, to the best of our knowledge, we are the first to investigate the multi-modal transformer network for this task.

\section{The Proposed Approach}

\subsection{Framework Overview}

As \cref{fig:2} illustrates, CAT-Det basically adopts a two-stream structure consisting of a Pointformer (PT) branch and an Imageformer (IT) branch, separately learning representations of LiDAR point-clouds and RGB images by exploring long-range intra-modal contexts. To complement the learning in each single modality, the Cross-Modal Transformer (CMT) module is employed to perform cross-modal feature interaction, followed by multi-modal feature aggregation with holistically learned fine-grained weights. The combination of PT, IT and CMT constitutes a novel transformer backbone. Meanwhile, a One-way Multi-modal Data Augmentation (OMDA) approach is developed to accomplish efficient data augmentation by a hierarchical contrastive learning at both the point-level and object-level, further facilitating the training of a strong deep transformer network for multi-modal 3D object detection.

\subsection{Two-stream Multi-modal Transformer}
\label{subsec:ptit}

Existing multi-modal 3D object detectors~\cite{huang2020epnet, yoo20203d} often adopt PointNet++/Sparse 3D CNN for point-clouds and 2D CNNs for images in representation learning. They mostly suffer from limited receptive fields, thus unable to fully explore global contextual information, which is important to detecting hard examples (\eg tiny objects). Recent work~\cite{devlin2018bert, dosovitskiy2020image} has proved the effectiveness of Transformer in modeling long-range dependencies. Despite its increasing prevalence, the transformer structure has not yet been investigated in multi-modal 3D object detection. This motivates us to make the first attempt in developing a deep multi-modal transformer backbone to capture richer global contexts for 3D object detection. 

To this end, we propose a novel multi-modal transformer network, consisting of the two-stream PT and IT branches connected by several CMTs as shown in \cref{fig:2}. Given a paired multi-modal input $\{\bm{P},I\}$, $PT(\cdot)$ and $IT(\cdot)$ learn representations for point-clouds $\bm{P}=\{\bm{p}_{1}, \bm{p}_{2}, \cdots, \bm{p}_{N}\}\in \mathbb{R}^{N\times 3}$ and images $I$, respectively. CMT performs cross-modal interaction and multi-modal aggregation at different levels.

\begin{figure}[t]
  \centering
   \includegraphics[width=0.99\linewidth]{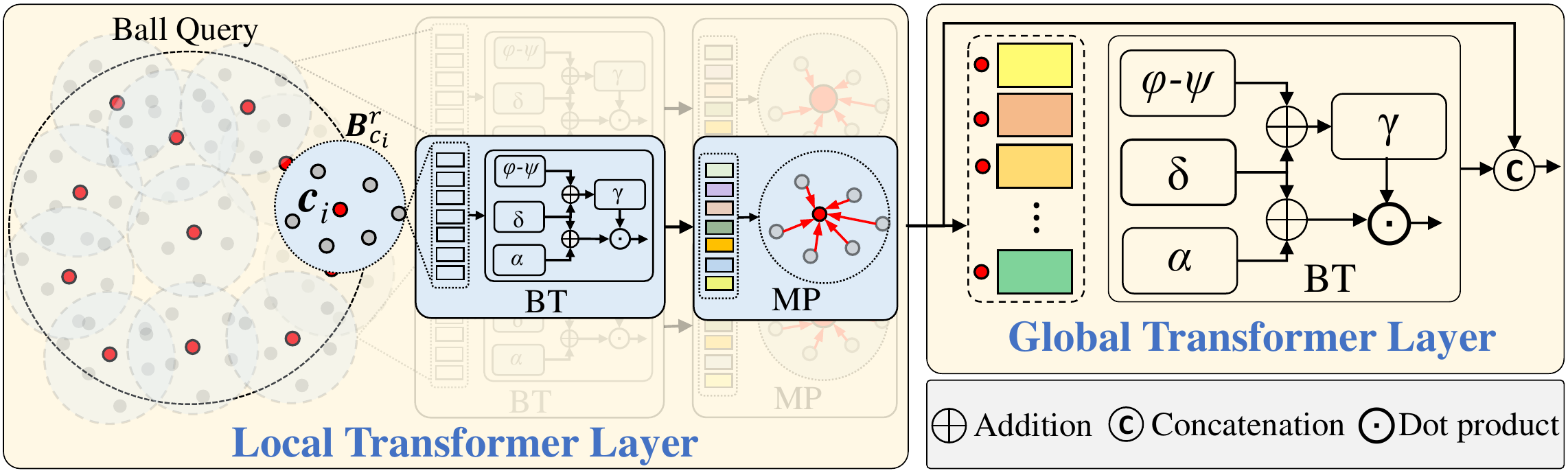}

   \caption{Point Transformer Block. By local and global combination, it captures both the dependencies from the adjacent regions and the whole scene, thus facilitating feature learning for 3D object detection. BT: basic transformer; MP: max-pooling.}
   \label{fig:3}
    \vspace{-0.4cm}
\end{figure}

\textbf{Pointformer.} Despite that a few recent attempts have investigated transformers for point-clouds~\cite{zhao2021point, guo2021pct}, most of them are specifically designed for classification, which only adopt a local transformer structure. However, global context information is vital to 3D detection, which cannot be fully recorded by local transformers. To address this issue, we propose a novel pointformer, composed of multiple stacked Point Transformer Blocks (PTB). As displayed in \cref{fig:3}, PTB consists of a local transformer layer and a global transformer layer. The local layer explores geometry structures of points within a neighborhood, and the global one encodes the holistic context at the scene level. By combining them, PTB captures the context information from the points of the nearby local regions as well as the full scene.

To be specific, the local transformer layer first applies the furthest point sampling on input point-clouds $\bm{P}$ to choose a subset $\bm{C}={\{\bm{c}_1,\bm{c}_2,\cdots,\bm{c}_{N'}\}}\subset \bm{P}$. Then, we conduct the ball query operation by taking each point $\bm{c}_i$ as the centroid, where $K$ points $\bm{B}_{\bm{c}_i}^r$ are selected within a ball centered at $\bm{c}_i$ with a radius $r$. The subset $\bm{B}_{\bm{c}_i}^r$ is further grouped and fed into a basic transformer block $BT(\cdot)$ for local information aggregation, which adopts the structure based on self-attention inspired by \cite{zhao2021point}. Given the input $\bm{B}_{\bm{c}_i}^r$, the output $\bm{y}_i=BT(\bm{B}_{\bm{c}_i}^r)$ is formulated as below:

\begin{equation}\label{eq:1}
 {\bm{y}_i} = \sum\limits_{{\bm{p}_j} \in \bm{B}_{\bm{c}_i}^r} {\rho (\gamma (\varphi ({\bm{p}_i}) - \psi ({\bm{p}_j}) + \delta ))}  \odot (\alpha ({\bm{p}_j}) + \delta),
\end{equation}
where the outputs of $\delta$, $\gamma$ and $\alpha$ are the encoded position, the self-attention and the transformed value, respectively. $\varphi$, $\psi$, and $\alpha$ are pointwise feature transformations, such as linear projections or MLPs. $\delta$ is a position encoding function defined as 
$\delta = \theta({\bm{p}}_i-\bm{p}_j)$.
Both the mapping function $\gamma$ and the encoding function $\theta$ are MLPs with two linear layers and one ReLU nonlinearity. $\rho$ is the Softmax function.

Albeit the exploration of long-range dependency in  $\bm{B}_{\bm{c}_i}^r$ by $BT(\cdot)$, the local transformer layer processes the point-cloud locally, unable to present the holistic context information. Therefore, we additionally employ a global transformer layer, which has a similar transformer structure as the local one, but takes all points $\bm{C}$ as input, instead of a local subset $\bm{B}_{\bm{c}_i}^r$. The features generated by the local and global transformer layer are concatenated to integrate both local and global contexts. 

Similar to PointNet++, we adopt a Feature Propagation (FP) layer after stacked PTBs for up-sampling.

\begin{figure}[!t]
  \centering
   \includegraphics[width=0.99\linewidth]{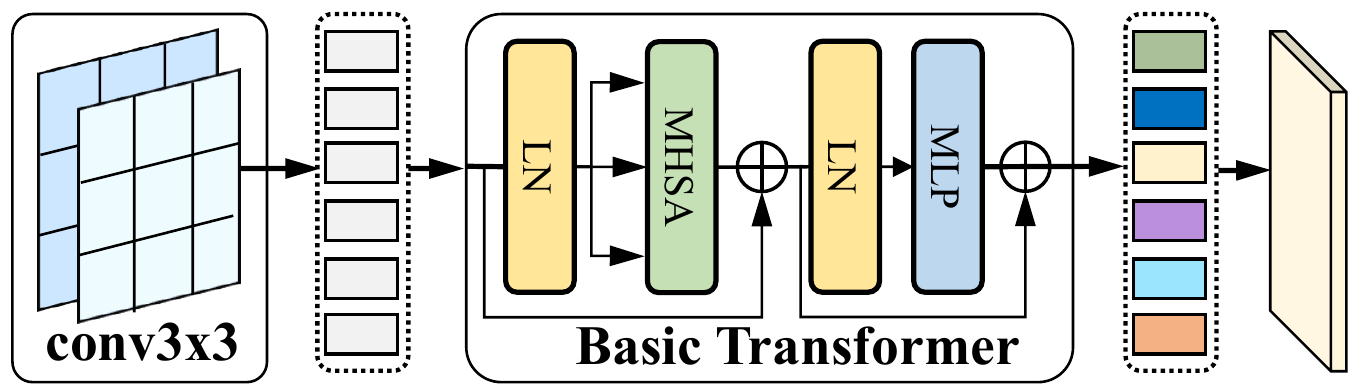}
   \caption{Image Transformer Block, which is a combination of a few convolutional layers and transformers.}
   \label{fig:4}
    \vspace{-0.4cm}
\end{figure}

\textbf{Imageformer.}  Vision Transformer (ViT)~\cite{dosovitskiy2020image} is the first work that adopts the transformer network in the visual domain, which employs the self-attention mechanism to build holistic dependencies among visual tokens. Since raw image patches are taken as tokens, it fails to encode local visual spatial information. Some recent studies~\cite{wu2021cvt,yuan2021incorporating} handle this problem by adding a few convolutional layers before the transformer layer, which is used as the basic transformer in our work. To align with Pointformer, we adopt similar structures by stacking several Image Transformer Blocks (ITB) as shown in \cref{fig:4}. Each ITB consists of two convolutional layers for local visual context encoding, and a successive basic multi-head transformer encoder~\cite{dosovitskiy2020image} for global context information exploration. Finally, ITB reshapes the transformed vector sequence into a 2D feature map for further processing. Following the stacked ITBs, an up-sampling (UP) layer is employed to recover the image resolution, generating feature maps with the same size as the original image.

\textbf{Cross-Modal Transformer.} PTB and ITB extensively explore contexts in the point-cloud $\bm{P}$ and the image $I$, respectively. However, as in \cref{fig:1}, the context in a single modality is probably incomplete due to noise, which can be complemented by that conveyed in the other modality. This motivates us to propose a module between PTB and ITB, to perform cross-modal information interaction and multi-modal feature aggregation. 

Suppose the features from PTB and ITB are $\bm{F}_{\bm{P}}$ and $\bm{F}_{I}$ respectively, where $\bm{F}_{\bm{P}}$ are representations of a set of down-sampled points $\bm{\hat{P}}\subset \bm{P}$. For each point $\bm{p}\in \bm{P}$, we project it to the corresponding pixel coordinate $\bm{p}'_{I}$ in $I$ by a function $f_{proj}(\cdot)$. For instance, in KITTI, $f_{proj}(\cdot)$ is formulated as:

\vspace{-2mm}
\begin{equation}
    \bm{p}'_{I} = f_{proj}(\bm{p})=C_{rect}\cdot R_{rect}\cdot T_{cam \leftarrow LiDAR}\cdot \bm{p},
    \label{eq:4}
\end{equation}
where $T_{cam \leftarrow LiDAR}$ is the transformation matrix from the coordinate of LiDAR to camera, $R_{rect}$ and $C_{rect}$ are the rectifying rotation and the calibration matrices of the camera, respectively. Based on $f_{proj}(\cdot)$, we convert the 3D coordinates $\bm{\hat{P}}$ to 2D pixels $\bm{\hat{P}}'_{I}=f_{proj}(\bm{\hat{P}})$, based on which we select the features from $\bm{F}_{I}$ at positions $\bm{\hat{P}}'_{I}$ and fetch a subset $\bm{F}_{I}'$ spatially aligned with $\bm{F}_{\bm{P}}$. In other words, $\bm{F}_{\bm{P}}$ and $\bm{F}'_I$ are point and imagery features for $\bm{\hat{P}}$, respectively.

\label{subsec:cmt}
\begin{figure}[!t]
  \centering
   \includegraphics[width=0.98\linewidth]{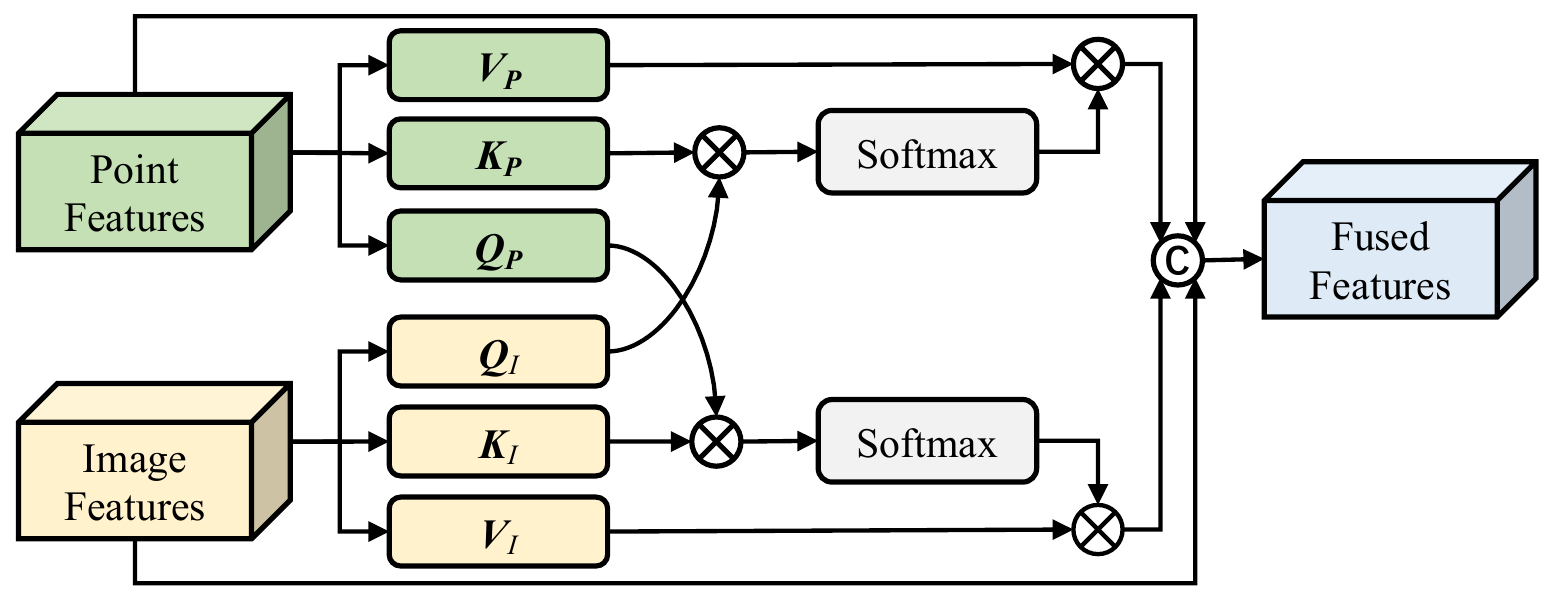}
\caption{Cross-Modal Transformer. It performs complementary feature enhancement by adaptively learning weights for different modalities through cross transformer.}
   \label{fig:5}
    \vspace{-0.4cm}
\end{figure}

Subsequently, CMT projects $\bm{F}_{\bm{P}}$ into the query $\bm{Q}_{\bm{P}}=\bm{F}_{\bm{P}}\cdot\bm{W}_{\bm{Q}}$, the key $\bm{K}_{\bm{P}}=\bm{F}_{\bm{P}}\cdot\bm{W}_{\bm{K}}$ and the value $\bm{V}_{\bm{P}}=\bm{F}_{\bm{P}}\cdot\bm{W}_{\bm{V}}$, where $\bm{W}_{\bm{Q}}$, $\bm{W}_{\bm{K}}$ and $\bm{W}_{\bm{V}}$ are learnable linear mappings. Similarly, the imagery feature $\bm{F}'_{I}$ is projected to $\bm{Q}_{I}$, $\bm{K}_{I}$ and $\bm{V}_{I}$. The context from the image modality is thus explored by attentive weights $\bm{A}_{\bm{P}\leftarrow I}=Softmax\left(\bm{Q}_{\bm{I}}\bm{K}_{\bm{P}}^{T}\right)$ and encoded into point features by $\bm{F}_{\bm{P}}^{cont}=\bm{A}_{\bm{P}\leftarrow I}\odot \bm{V}_{\bm{P}}$. Similarly, the context from the point modality can be explored and encoded into imagery features by $\bm{F}_{I}^{cont}=Softmax\left(\bm{Q}_{\bm{P}}\bm{K}_{I}^{T}\right)\odot \bm{V}_{I}$.

The original multi-modal features $\bm{F}_{\bm{P}}$/$\bm{F}_{I}$ and the features $\bm{F}_{\bm{P}}^{cont}$/$\bm{F}_{I}^{cont}$ with cross-modal interactions are aggregated by $\bm{F}_{\bm{P}}:=\bm{F}_{\bm{P}}\oplus\bm{F}_{I}\oplus\bm{F}_{\bm{P}}^{cont}\oplus\bm{F}_{I}^{cont}$ as new point features, where $\oplus$ stands for concatenation.

\subsection{One-way Multi-modal Data Augmentation}
\label{subsec:omda}
Data augmentation has proved effective for object detection, which however is mostly applied within a single modality and rarely considered in the multi-modal scenario. Due to the heterogeneity between point-clouds and images, it is generally difficult to synchronize the augmenting operations across modalities, leading to severe cross-modal misalignment. Recently, ~\cite{wang2021pointaugmenting} presents a complex approach to generate paired data, but the pipeline on images is cumbersome and easy to incur noise. Instead, we propose a novel One-way Multi-modal Data Augmentation (OMDA) approach, which performs augmentation on point-clouds only, and efficiently extends it to multiple modalities by contrastive learning. 

\begin{figure}[t]
  \centering
   \includegraphics[width=0.95\linewidth]{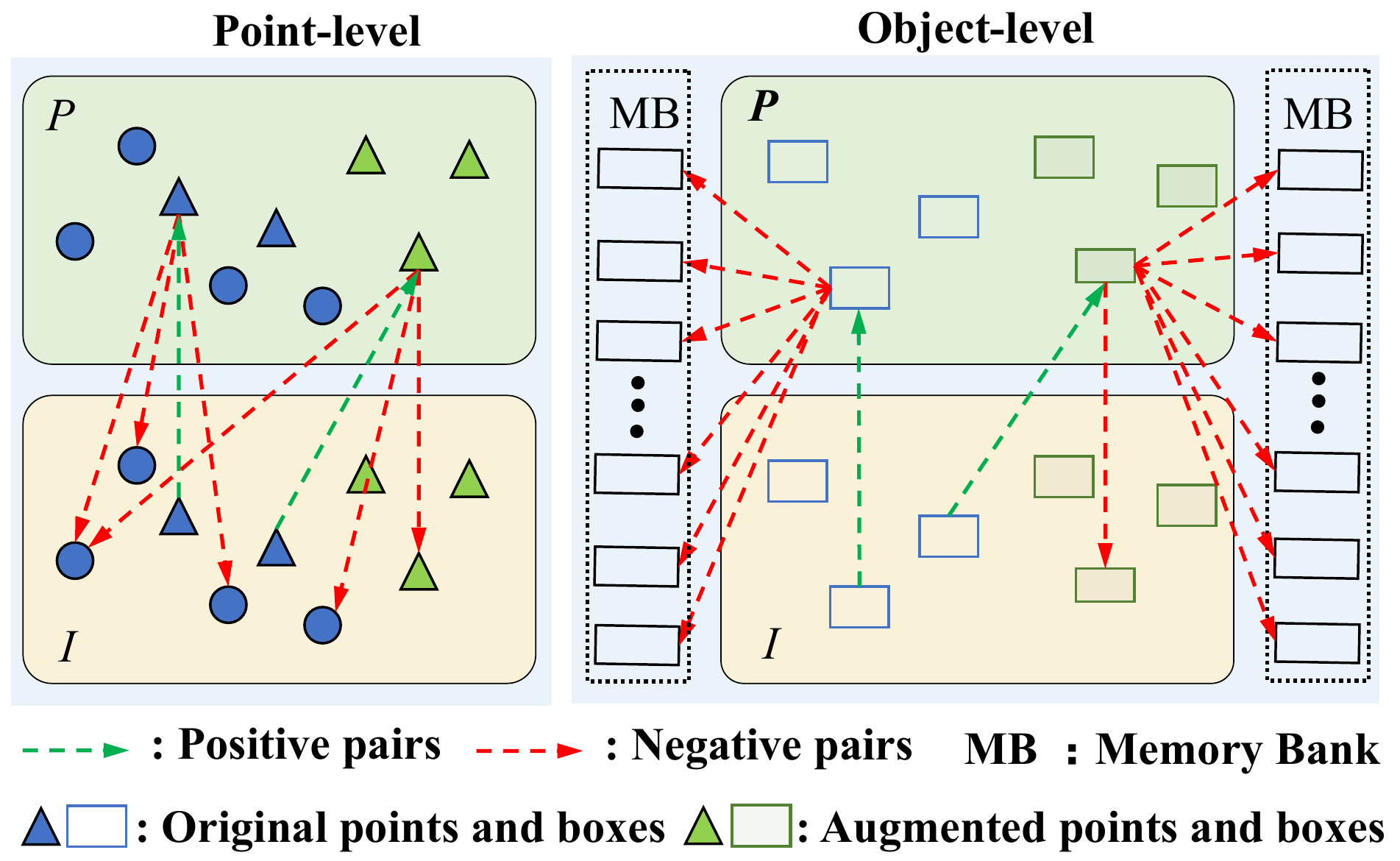}

   \caption{Positive/negative pair selection for contrastive learning.}
   \label{fig:6}
   \vspace{-0.4cm}
\end{figure}

The basic idea behind OMDA is two-fold: (1) High-quality image augmentation is generally much more complex and difficult than that on point-clouds and it is thus expected to only augment LiDAR data and then make a light-weight multi-modal extension. (2) One-way augmentation as in (1) may bring in severe cross-modal misalignment. Inspired by the recent success of contrastive learning in self-supervised models \cite{he2020momentum, chen2020simple} and cross-modal semantic alignment \cite{liu2020p4contrast, liu2021hit}, we elaborately design a contrastive learning scheme to address such misalignment across modalities.

Specifically, OMDA adopts GT-Paste~\cite{yan2018second}, widely used for LiDAR-only methods, to augment a given point-cloud by pasting extra 3D objects from other LiDAR frames without spatial collision. As there lacks the images corresponding to the augmented point-clouds, cross-modal data misalignment occurs, probably deteriorating multi-modal interaction (\eg CMT), which implicitly assumes that point-cloud/image pairs are well-aligned. Thus, we perform contrastive learning among the raw point-clouds $\bm{P}$, the corresponding images $I$, and the augmented point-clouds $\bm{P}_{aug}$ at both point- and object-level in a hierarchical manner. 

\textbf{Point-level Contrastive Augmentation.} To preserve the supervision of raw data pair, we firstly construct cross-modal positive/negative point pairs from $(\bm{P},I)$ for constrative learning. Given a point $\bm{p}\in \bm{P}$, we fetch its corresponding 2D pixel coordinate $\bm{p}'_{I}$ by $\bm{p}'_{I}=f_{proj}(\bm{p})$ as in \cref{eq:4}. Since $\bm{P}$ and $I$ are well aligned, $(\bm{p}, \bm{p}'_{I})$ indicates the 3D/2D position of the same object, thus naturally forming the positive pair. To construct negative pairs, we select the 3D points $\mathcal{N}_{\bm{p}}\subset \bm{P}$ that belong to different object classes from $\bm{p}$, \emph{e.g.} points that with confidence scores (predicted by the segmentation head) less than a threshold $t$. The negative pairs are chosen as $\{(\bm{p},\bm{q}'_{I})|\bm{q}'_{I}\in \mathcal{N}'_{\bm{p},I}\}$,
where $\mathcal{N}'_{\bm{p},I}$ is the set of 2D coordinates of $\mathcal{N}_{\bm{p}}$, \ie $\mathcal{N}'_{\bm{p},I}=f_{proj}(\mathcal{N}_{\bm{p}})$. 

Afterwards, we construct positive/negative point pairs from the augmented point-cloud $\bm{P}_{aug}$ and the unpaired image $I$ as in \cref{fig:6}. Suppose $\bm{p}_{aug}\in \bm{P}_{aug}$ is a point from the pasted virtual object $O_{vir}$. Since GP-paste avoids spatial overlap with existing objects when pasting $O_{vir}$ on $\bm{P}$, no objects lie at the position $f_{proj}(\bm{p}_{aug})$ in $I$. It means that $(\bm{p}_{aug}, f_{proj}(\bm{p}_{aug}))$ is definitely unpaired, thus forming a negative pair. To collect positive pairs w.r.t. $\bm{p}_{aug}$, we select the 3D point $\bm{\hat{p}} \in \bm{P}$ that most likely belongs to the same class as $\bm{p}_{aug}$, \emph{e.g.} the point with the highest confidence score predicted by the segmentation head. Thus, $(\bm{p}_{aug}, f_{proj}(\bm{\hat{p}}))$ is chosen as the positive pair.

Based on the selected positive/negative pairs $\mathcal{S}_{+}$/$\mathcal{S}_{-}$, we formulate the following point-level contrastive loss as:
\begin{equation}
    \mathcal{L}_{cl-p} = -\sum_{(i,j) \in \mathcal{S}_{+}}\log\frac{\exp(\bm{f}_i^{T}\cdot\bm{f}_j/\tau)}{\sum_{(i,k) \in \mathcal{S}_{-}}\exp(\bm{f}_i^{T}\cdot\bm{f}_k/\tau)},
    \label{eq:9}
\end{equation}
where $\bm{f}_{i}$ is the feature of the $i$-th point in $\bm{P}$, $\bm{f}_{j}$ is the feature at location $j$ of image $I$, and $\tau$ is the scaling factor. It can be observed that the correlation between the paired cross-modal features is increased, while being decreased for unpaired features by minimizing $\mathcal{L}_{cl-p}$, thus alleviating cross-modal misalignment of the augmented data.  

\textbf{Object-level Contrastive Augmentation.} Contrastive learning at the point-level offers fine-grained point-wise semantic alignment, while detectors focus on regions. To alleviate regional semantic alignment, we additionally perform object-level contrastive learning on augmented data. 

Similar to point-level contrastive learning, we first construct positive/negative pairs of objects across modalities. As in \cref{fig:6}, the object $O$ from the raw point-cloud $\bm{P}$ and its paired object $O_{I}'$ in the aligned image $I$ naturally constitute a positive pair. The pasted virtual object $O_{vir}$, belonging to the same class as $O$, also forms a positive pair with $O_{I}'$. Due to the class imbalance frequently occurring in labeled data, $\bm{P}$ and $I$ probably only contain objects from one class. Although we can simply select background areas to form the negative pair, foreground areas belonging to objects from different classes are more desirable, as they provide stronger supervision. Inspired by~\cite{he2020momentum}, we employ a memory bank to generate more precise and discriminative representations for negative pair selection. As in \cref{fig:2}, the memory bank adopts an encoder $E(\cdot)$ with the same structure as the two-stream multi-modal transformer, and maintains two queues of features denoted by $\mathcal{Q}_{\bm{P}}$ and $\mathcal{Q}_{I}$. $\mathcal{Q}_{\bm{P}}$ contains point features of objects from all classes, and $\mathcal{Q}_{I}$ contains imagery features. For $O$ or $O'_{vir}$, we select the elements in $\mathcal{Q}_{\bm{P}}$ and $\mathcal{Q}_{I}$ from distinct classes to collect negative pairs.

With positive and negative object pairs $\mathcal{O}_{+}$/$\mathcal{O}_{-}$, object-level contrastive learning is applied by minimizing the loss:

\begin{equation}
    \mathcal{L}_{cl-o} = -\sum_{(i,j) \in \mathcal{O}_{+}} \log\frac{\exp(\bm{g}_i^{T}\cdot\bm{g}_j/\tau)}{\sum_{(i,k) \in \mathcal{O}_{-}}\exp(\bm{g}_i^{T}\cdot\bm{g}_k/\tau)},
    \label{eq:oc}
\end{equation}
where $\bm{g}$ is the object-level representation by aggregating feature vectors in the object bounding box via max-pooling or is directly fetched from the memory banks $\mathcal{Q}_{\bm{P}}$/$\mathcal{Q}_{I}$.

As in \cite{he2020momentum}, we employ the momentum update mechanism to optimize the encoder $E(\cdot)$ instead of gradient update, in order to strengthen the stability of features in the memory bank. Refer to \cite{he2020momentum} for more details about the optimization on $E(\cdot)$ and memory banks $\mathcal{Q}_{\bm{P}}$/$\mathcal{Q}_{I}$.

\begin{table*}[t]
\centering
\resizebox{2.0\columnwidth}{!}{
\begin{tabular}{l|c|cccc|cccc|cccc|c}
\hline
\multirow{2}{*}{Method} &\multirow{2}{*}{Modality} &\multicolumn{4}{|c|}{Car (\%)} &\multicolumn{4}{|c|}{Pedestrian (\%)} & \multicolumn{4}{|c|}{Cyclist (\%)} &\multirow{2}{*}{mAP (\%)}\\
\cline{3-14}
 &&Easy &Mod. &Hard &mAP&Easy &Mod. &Hard&mAP &Easy &Mod. &Hard&mAP\\
\hline
VoxelNet~\cite{zhou2018voxelnet} &L &77.47	&65.11	&57.73&66.77	&39.48	&33.69	&31.51&34.89	&61.22	&48.36	&44.37&51.32	&50.99\\
PointRCNN~\cite{shi2019pointrcnn} &L	&86.96	&75.64	&70.70&77.77	&47.98	&39.37	&36.01&41.12	&74.96	&58.82	&52.53&62.10	&60.33\\
PointPillars~\cite{lang2019pointpillars} &L &82.58	&74.31	&68.99&75.29	&51.45	&41.92	&38.89&44.09	&77.10	&58.65	&51.92&62.56	&60.65\\
TANet~\cite{liu2020tanet} &L &84.39	&75.94	&68.82&76.38	&53.72	&44.34	&40.49&46.18	&75.70	&59.44	&52.53&62.56	&61.71\\
STD~\cite{yang2019std}	&L	&87.95	&79.71	&75.09&80.92	&53.29	&42.47	&38.35&44.70	&78.69 &61.59	&55.30&65.19	&63.60\\
Part-A$^2$~\cite{shi2020points}	&L	&87.81	&78.49	&73.51&79.94	&53.10	&43.35	&40.06&45.50	&79.17	&63.52	&56.93&66.54	&63.99\\
PV-RCNN~\cite{shi2020pv} &L &90.25	&81.43	&76.82&82.83	&52.17 	&43.29	&40.29&45.25	&78.60	&63.71	&57.65&66.65	&64.91\\
3DSSD~\cite{yang20203dssd} &L &88.36	&79.57	&74.55&80.83	&\textbf{54.64} 	&44.27	&40.23&46.38	&82.48	&64.10	&56.90&67.82	&65.01\\
HotSpotNet~\cite{chen2020object} &L &87.60	&78.31	&73.34&79.75	&53.10 	&45.37	&41.47&46.65	&82.59	&65.95	&59.00&69.18	&65.19\\
SA-SSD~\cite{he2020structure} &L &88.75	&79.79	&74.16&80.90	&-- 	&--	&--&--	&--	&--	&--&--	&--\\
SE-SSD~\cite{zheng2021se} &L &\textbf{91.49}	&\textbf{82.54}	&\textbf{77.15}&\textbf{83.73}	&-- 	&--	&--&--	&--	&--	&--&--	&--\\
\hline
PointTransformer~\cite{pan20213d} &L &87.13	&77.06	&69.25&77.81	&50.67 	&42.43	&39.60&44.23	&75.01	&59.80	&53.99&62.93	&61.66\\
M3DETR~\cite{guan2021m3detr} &L &90.28	&81.73	&76.96&82.99	&45.70 	&39.94	&37.66&41.10	&\textbf{83.83}	&66.74	&59.03&69.87	&64.65\\
\hline
MV3D~\cite{chen2017multi} &L+R	&74.97	&63.63	&54.00&64.20	&--	&--	&--&--	&--	&--	&--&--	&--\\
ContFuse~\cite{liang2018deep} &L+R	&83.68	&68.78	&61.67&71.38	&--	&--	&--&--	&--	&--	&--	&--&--\\
MMF~\cite{liang2019multi} &L+R &88.40 &77.43 &70.22&78.68	&--	&--	&--&--	&--	&--	&--&--	&--\\
PI-RCNN~\cite{xie2020pi} &L+R &84.37 &74.82 &70.03&76.41&--	&--	&--&--	&--&--	&--	&--&--\\
EPNet~\cite{huang2020epnet} &L+R &89.81 &79.28 &74.59	&81.23	&--	&--&--	&--&--	&--&--	&--	&--\\
3D-CVF~\cite{yoo20203d} &L+R &89.20 &80.05 &73.11&80.79	&--	&--	&--&--	&--	&--	&--&--	&--\\
CLOCs~\cite{pang2020clocs} &L+R &88.94 &80.67 &\textbf{77.15}&82.25	&--	&--	&--	&--&--	&--	&--	&--&--\\
AVOD-FPN~\cite{ku2018joint} &L+R	&83.07	&71.76	&65.73&73.52	&50.46	&42.27	&39.04&43.92	&63.76	&50.55	&44.93&53.08	&56.84\\
F-PointNet~\cite{qi2018frustum} &L+R &82.19 &69.79	&60.59&70.86	&50.53	&42.15	&38.08&43.59	&72.27	&56.12	&49.01&59.13	&57.86\\
PointPainting~\cite{vora2020pointpainting} &L+R &82.11 &71.70	&67.08&73.63	&50.32	&40.97	&37.84&43.05	&77.63	&63.78	&55.89&65.77	&60.82\\
F-ConvNet~\cite{wang2019frustum} &L+R &87.36 &76.39	&66.69&76.81	&52.16	&43.38	&38.80&44.78	&81.98	&65.07	&56.54&67.86	&63.15\\
\hline
CAT-Det (\textbf{Ours}) &L+R  &89.87 &81.32 &76.68&82.62	&54.26	&\textbf{45.44}	&\textbf{41.94}&\textbf{47.21}	&83.68	&\textbf{68.81}	&\textbf{61.45}&\textbf{71.31}	&\textbf{67.05}\\
\hline
\end{tabular}
}
\caption{Comparison with state-of-the-art approaches on the KITTI test split. `-' indicates that either the result is not reported or the source code is not publicly available. `L' and `R' stand for the LiDAR and RGB modalities, respectively. Best in bold.}
\label{table:1}
\end{table*}

\textbf{Overall optimization.} Besides the contrastive learning losses $\mathcal{L}_{cl-p}$ and $\mathcal{L}_{cl-o}$, we also utilize the conventional detection losses $\mathcal{L}_{rpn}$ and $\mathcal{L}_{rcnn}$ as in \cite{shi2019pointrcnn}. The total loss for optimizing the overall transformer network is formulated as $\mathcal{L}_{tot}=\mathcal{L}_{rpn}+\mathcal{L}_{rcnn}+\lambda\cdot(\mathcal{L}_{cl-p}+\mathcal{L}_{cl-o})$, where $\lambda$ is the trade-off parameter, empirically set as 0.15 by default.

\section{Experiments}

We evaluate CAT-Det on the widely used KITTI benchmark~\cite{geiger2012we}, and for fair comparison, we adopt the same protocol as in~\cite{chen2017multi, shi2019pointrcnn}, separating original training data into a training set and a validation set. The Average Precision (AP) is used as the metric, and the IoU thresholds are set to 0.7, 0.5, and 0.5 for car, pedestrian and cyclist, as officially specified. APs are computed by recalling 11 and 40 positions on the val and test splits respectively.

\begin{table*}[t]
\centering
\resizebox{2.0\columnwidth}{!}{
\begin{tabular}{l|c|cccc|cccc|cccc|c}
\hline
\multirow{2}{*}{Method} &\multirow{2}{*}{Modality} &\multicolumn{4}{|c|}{Car (\%)} &\multicolumn{4}{|c|}{Pedestrian (\%)} & \multicolumn{4}{|c|}{Cyclist (\%)} &\multirow{2}{*}{mAP (\%)}\\
\cline{3-14}
 &&Easy &Mod. &Hard &mAP&Easy &Mod. &Hard&mAP &Easy &Mod. &Hard&mAP\\
\hline
PointPillars~\cite{lang2019pointpillars} &L &86.46	&77.28	&74.65&79.46	&57.75	&52.29	&47.90&52.65	&80.05	&62.68	&59.70&67.48	&66.53\\
SECOND~\cite{yan2018second} &L &88.61	&78.62	&77.22&81.48	&56.55	&52.98	&47.73&52.42	&80.58	&67.15	&63.10&70.28	&68.06\\
3DSSD~\cite{yang20203dssd} &L &88.55	&78.45	&77.30&81.43	&58.18 	&54.31	&49.56&54.02	&86.25	&70.48	&65.32&74.02	&69.82\\
PointRCNN~\cite{shi2019pointrcnn} &L	&88.72	&78.61	&77.82&81.72	&62.72	&53.85	&50.24&55.60	&86.84	&71.62	&65.59&74.68	&70.67\\
PV-RCNN~\cite{shi2020pv} &L &89.03	&83.24	&78.59&83.62	&63.71 	&57.37	&52.84&57.97	&86.06	&69.48	&64.50&73.35	&71.64\\
Part-A$^2$~\cite{shi2020points}	&L	&89.55	&79.40	&78.84&82.60	&65.68	&60.05	&55.44&60.39	&85.50	&69.90	&65.48&73.63	&72.20\\
SE-SSD~\cite{zheng2021se} &L &\textbf{90.21}	&\textbf{86.25}	&\textbf{79.22}&\textbf{85.23}	&-- 	&--	&--&--	&--	&--	&--&--	&--\\
\hline
MV3D~\cite{chen2017multi} &L+R	&71.29	&62.68	&56.56&63.51	&--	&--	&--&--	&--	&--	&--&--	&--\\
3D-CVF~\cite{yoo20203d} &L+R &89.67 &79.88 &78.47&82.67	&--	&--	&--&--	&--	&--	&--&--	&--\\
AVOD-FPN~\cite{ku2018joint} &L+R	&84.41	&74.44	&68.65&75.83	&--	&58.80	&--&--	&--	&49.70	&--&--	&--\\
PointFusion~\cite{xu2018pointfusion} &L+R	&77.92	&63.00	&53.27&64.73	&33.36	&28.04	&23.38&28.26	&49.34	&29.42	&26.98	&35.25&42.75\\
F-PointNet~\cite{qi2018frustum} &L+R &83.76 &70.92	&63.65&72.78	&70.00	&61.32	&53.59&61.64	&77.15	&56.49	&53.37&62.34	&65.58\\
SIFRNet~\cite{zhao20193d} &L+R &85.62 &72.05	&64.19&73.95	&69.35	&60.85	&52.95&61.05	&80.87	&60.34	&56.69&65.97	&66.99\\
CLOCs~\cite{pang2020clocs} &L+R &89.49 &79.31 &77.36&82.05	&62.88	&56.20	&50.10	&56.39&87.57	&67.92	&63.67	&73.05&70.50\\
EPNet~\cite{huang2020epnet} &L+R &88.76 &78.65 &78.32	&81.91	&66.74	&59.29&54.82	&60.28&83.88	&65.50&62.70	&70.69	&70.96\\
\hline
CAT-Det (\textbf{Ours}) &L+R  &90.12 &81.46 &79.15&83.58	&\textbf{74.08}	&\textbf{66.35}	&\textbf{58.92}&\textbf{66.45}	&\textbf{87.64}	&\textbf{72.82}	&\textbf{68.20}&\textbf{76.22}	&\textbf{75.42}\\
\hline
\end{tabular}
}
\caption{Comparison with state-of-the-art approaches on the KITTI validation split. Best in bold.}
\label{table:2}
\vspace{-0.4cm}
\end{table*}
\subsection{Implementation Details}

Both the point-clouds and images are used in training and testing. The range of point-clouds is constrained to (0, 70.4), (-40, 40) and (-3, 1) along the X, Y and Z axis, respectively, which is further down-sampled to 16,384 points as input, and the resolution of images is 1280$\times $384. In the PT branch, there are four stacked PTBs, with the numbers of sampled points set to 4,096, 1,024, 256, and 64, respectively, and four FP layers which up-sample the point-cloud back to the original size with a stride of 4. Similarly, in the IT branch, there are four cascaded ITBs followed by four UP layers for parallel transposed convolutions with strides 2, 4, 8, and 16. In our memory bank, the sample size of each category is 1,024. We adopt the ADAM optimizer and the cosine annealing learning rate schedule with an initial value at 0.002. The batch size and the maximal number of learning epochs are set to 16 and 80, respectively. All the experiments are conducted on 8 GTX 1080Ti GPUs. 

\subsection{Comparison with State-of-the-Arts}
We compare CAT-Det to the following categories of methods, including (1) LiDAR-only with non-transformer structures \cite{zhou2018voxelnet, yan2018second, shi2019pointrcnn, lang2019pointpillars, liu2020tanet, yang2019std, shi2020points, shi2020pv, yang20203dssd, chen2020object, he2020structure, zheng2021se}; (2) LiDAR-only with transformer structures \cite{pan20213d, guan2021m3detr}; (3) Multi-modal (LiDAR+RGB) \cite{chen2017multi, liang2018deep, xu2018pointfusion, liang2019multi, zhao20193d, xie2020pi, huang2020epnet, yoo20203d, pang2020clocs, ku2018joint, qi2018frustum, vora2020pointpainting, wang2019frustum}.

\begin{figure*}[t]
\centering
\includegraphics[width=0.95\textwidth]{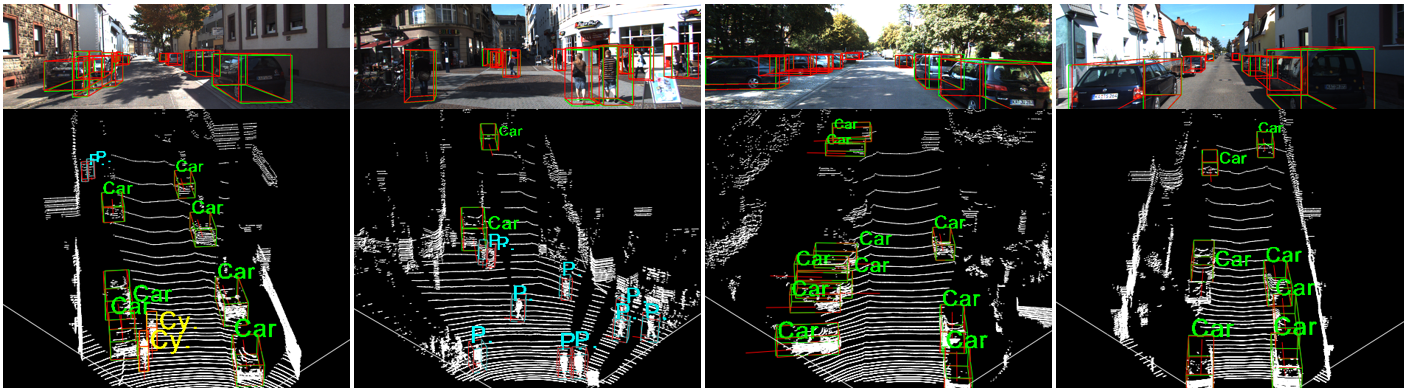} 
\caption{Visualized results by CAT-Det on KITTI val split. Red/green rectangles indicate predicted/GT bounding boxes.}
\label{fig:7}
\vspace{-0.2cm}
\end{figure*}

Table~\ref{table:1} summarizes the official results on the test set. As Table~\ref{table:1} displays, the LiDAR-only methods outperform the existing multi-modal counterparts at most cases, indicating that modeling multi-modal data indeed remains a challenging task, despite more information available. PointTransformer and M3DETR both adopt the transformer structure; however, their performance is not as good as that of the non-transformer counterparts such as\cite{yang20203dssd} and~\cite{chen2020object}. By virtue of the specially designed transformer structure and the effective one-way multi-modal data augmentation scheme, our method reaches a newly the-state-of-art score on KITTI. It is worth noting that CAT-Det is the first multi-modal solution that surpasses the LiDAR-only
ones with a large margin, \ie. a gain of 1.86\% over the second best HotSpotNet, suggesting the potential of multi-modal data for further improvement. Besides, CAT-Det ranks the first in many cases, particularly for the challenging classes of pedestrian and cyclist. In these two classes, there are many small or partial instances, and more context information with large receptive fields is thus required, which is fully explored by the transformer structures of CAT-Det. 

We also evaluate the methods on KITTI val. As Table \ref{table:2} shows, CAT-Det again achieves the best mAP, 3.22\% higher than the second best Part-A$^2$. The decent results on hard examples visualized in \cref{fig:7} highlight the advantage of CAT-Det for this issue.

\subsection{Ablation Study}

\noindent \textbf{On TPI/CMT/OMDA.} We individually evaluate the contributions of TPI, CMT and OMDA in CAT-Det. PointNet++ and 2D CNN with concatenation based fusion is selected as the baseline, whose mAP is 68.65\%. After replacing PointNet++ and 2D CNN with TPI while maintaining the concatenation operation, the performance is boosted by 1.79\%, and it is further improved by 1.45\% through incorporating CMT. We attribute this improvement to the specifically designed network structures, which extensively capture intra-modal and cross-modal global contexts. Finally, we apply OMDA and obtain the full model of CAT-Det, which gains 3.53\% in mAP, validating its effectiveness.  

\begin{table}[t]
\centering
\resizebox{0.95\columnwidth}{!}{
\begin{tabular}{ccc|ccc|c}
\hline
\multicolumn{3}{c|}{Component} &\multicolumn{4}{|c}{3D Object Detection (\%)}\\
\hline
TPI	&CMT &OMDA	&Car	&Ped.	&Cyc.	&mAP	\\
\hline
  & & &80.41	&59.27	&66.28	&68.65	\\
\textbf{$\checkmark$} &	&	&81.12	&61.26	&68.94	&70.44\\
 \textbf{$\checkmark$}&	\textbf{$\checkmark$}&	&81.73	&63.30	&70.63	&71.89\\
\textbf{$\checkmark$} &\textbf{$\checkmark$} &\textbf{$\checkmark$}&\textbf{83.58}	&\textbf{66.45}	&\textbf{76.22}	&\textbf{75.42}\\
\hline
\end{tabular}
}
\caption{Contributions of TPI/CMT/OMDA to CAT-Det.}
\label{table:3}
\end{table}

\begin{table}[t]
\centering
\resizebox{0.95\columnwidth}{!}{
\begin{tabular}{cccc|ccc|c}
\hline
\multicolumn{4}{c|}{Module} &\multicolumn{4}{|c}{3D Object Detection (\%)}\\
\hline
Concat.	&AF &CMT$_S$ &CMT	&Car	&Ped.	&Cyc.	&mAP	\\
\hline
\textbf{$\checkmark$} & & & &81.12	&61.26	&68.94	&70.44	\\
&\textbf{$\checkmark$} &	&	&81.31	&61.74	&69.45	&70.83\\
 &&	\textbf{$\checkmark$}&	&81.57	&62.68	&69.99	&71.41\\
&& &\textbf{$\checkmark$}&\textbf{81.73}	&\textbf{63.30}	&\textbf{70.63}	&\textbf{71.89}\\
\hline
\end{tabular}
}
\caption{Results of different fusion schemes in the CMT module.}
\label{table:4}
\vspace{-0.4cm}
\end{table}

\noindent \textbf{On Fusion in CMT.} To analyze CMT in integrating multi-modal features, the widely used alternatives are applied for comparison, including concatenation (Concat.)~\cite{chen2017multi, ku2018joint} and attention-based fusion (AF) ~\cite{yoo20203d, huang2020epnet}. A variant of CMT (denoted as CMT$_S$) is also considered, where only the cross-modal features are concatenated as $\bm{F}_{\bm{P}}^{cont}\oplus \bm{F}_{I}^{cont}$ as the output, without single-modal features $\bm{F}_{\bm{P}}$ and $\bm{F}_{I}$. As in Table \ref{table:4}, CMT clearly outperforms Concat. and AF only with cross-modal features (CMT$_S$), and further boosts the accuracy by preserving single-modal features. It proves the advantage of CMT in encoding cross-modal global contexts.

\noindent \textbf{On Components in OMDA.} To validate point-level and object-level contrastive learning, three versions of OMDA are adopted: (1) point-level augmentation only (\emph{CA-P}); (2) object-level augmentation with background to select negative pairs (\emph{CA-O-BG}); (3) object-level augmentation with memory bank to select negative pairs (\emph{CA-O-MB}). As Table \ref{table:5} displays, a gain of 2.65\% is achieved when only applying point-level contrastive augmentation, clearly revealing the superiority of OMDA. The performance is further boosted using a simple version of object-level contrastive augmentation, \ie \emph{CA-O-BG}, validating the necessity of augmentation at the object-level. By expanding object-level negative sample pairs via memory bank, the score is the best.

\begin{table}[t]
\centering
\resizebox{0.95\columnwidth}{!}{
\begin{tabular}{ccc|ccc|c}
\hline
\multicolumn{3}{c|}{Method} &\multicolumn{4}{|c}{3D Object Detection (\%)}\\
\hline
CA-P	& CA-O-BG & CA-O-MB	&Car	&Ped.	&Cyc.	&mAP	\\
\hline
  & & &81.73	&63.30	&70.63	&71.89	\\
\textbf{$\checkmark$} &	&	&82.95	&66.13	&74.55	&74.54\\
 \textbf{$\checkmark$}&	\textbf{$\checkmark$}&	&83.37	&66.27	&75.36	&75.00\\
\textbf{$\checkmark$} & &\textbf{$\checkmark$}&\textbf{83.58}	&\textbf{66.45}	&\textbf{76.22}	&\textbf{75.42}\\
\hline
\end{tabular}
}
\caption{Results of different components in OMDA.}
\label{table:5}
\vspace{-0.4cm}
\end{table}

\section{Conclusion}
This paper proposes a novel framework, namely Contrastively Augmented Transformer for multi-modal 3D object Detection (CAT-Det). It aims to solve the problems of insufficient multi-modal fusion and lack of effective multi-modal data augmentation. To this end, we propose a multi-modal transformer to extensively encode the intra-modal and cross-modal long-range context information. Furthermore, we propose a one-way data augmentation via hierarchical contrastive learning, remarkably improving the accuracy only by augmenting point-clouds, thus free from complex generation of paired samples of the two modalities. Extensive experiments are conducted on KITTI, and CAT-Det reaches a newly state-of-the-art.

\section*{Acknowledgment}
This work is partly supported by the National Natural Science Foundation of China (No. 62022011), the Research Program of State Key Laboratory of Software Development Environment (SKLSDE-2021ZX-04), and the Fundamental Research Funds for the Central Universities.

{\small
\bibliographystyle{ieee_fullname}
\bibliography{egbib}
}

\clearpage
\setcounter{table}{0}
\setcounter{figure}{0}
\setcounter{section}{0}
\renewcommand\thesection{\Alph{section}}
\renewcommand\thefigure{\Alph{figure}}
\renewcommand\thetable{\Alph{table}}

\noindent \textbf{\Large Supplementary Material} \\

\begin{figure*}[t]
\centering
\includegraphics[width=0.99\textwidth]{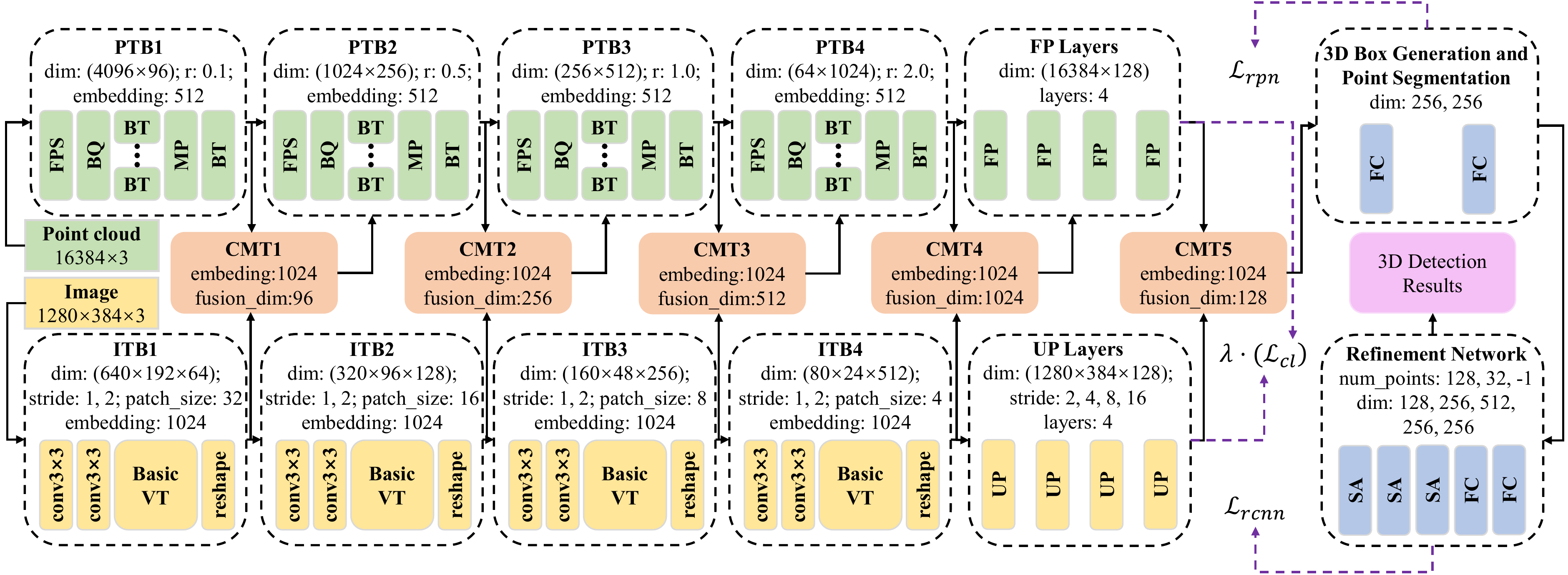} 
\caption{Detailed network architecture of the proposed CAT-Det framework. PTB: the point transformer block; ITB: the image transformer block; CMT: the cross-modal transformer; FPS: the farthest point sampling; BQ: the ball query; BT: the basic point transformer; VT: the basic vision transformer; MP: the max-pooling; UP: the up-sampling layer; FP: the feature propagation layer; SA: the set abstraction layer; and FC: the fully connected layer.}
\label{fig:A}
\end{figure*}

In this supplementary document, we provide more implementation details of the proposed CAT-Det framework in~\cref{sec:A} as well as more ablation results on the main components of our approach in~\cref{sec:B} including: the CMT module, the memory bank size, the trade-off parameter $\lambda$ balancing the detection loss $(\mathcal{L}_{rpn}+\mathcal{L}_{rcnn})$ and contrastive learning loss $(\mathcal{L}_{cl-p}+\mathcal{L}_{cl-o})$, the GT-Paste and the ITB and PTB blocks. The runtime and complexity analysis of CAT-Det are also reported  in~\cref{sec:B}. In addition, we demonstrate more visualization results of CAT-Det on KITTI val split in~\cref{sec:C}, and show detailed scores on the official KITTI test leaderboard in \cref{sec:D}.

\section{More Implementation Details}
\label{sec:A}
In this section, we describe the details of the major parts in CAT-Det (as depicted in \cref{fig:2} in the main paper), including pointformer, imageformer, one-way multi-modal data augmentation (OMDA), and 3D box generation and point segmentation, together with the training losses.

The detailed network architecture is illustrated in Fig.~\ref{fig:A}, and we elaborate each critical component as follows.

As displayed in \cref{fig:2} and \cref{fig:3} from the main paper, \textbf{Pointformer} consists of four point transformer blocks (PTBs), where the radii for ball query in PTBs are set to $[0.1, 0.5, 1.0, 2.0]$ and the channel sizes are fixed as $[96, 256, 512, 1024]$, respectively. The linear projection dimension in basic point transformer (BT) is set to 512. Similar to PointNet++, four feature propagation (FP) layers are adopted after stacked PTBs for up-sampling the point-cloud back to the original size with a stride of 4.

As shown in \cref{fig:4}, \textbf{Imageformer} is composed of four image transformer blocks (ITBs), where the channel sizes are $[64, 128, 256, 512]$, respectively. For basic transformer, we use 4 self attention heads, of which the linear projection dimensions are fixed as 1024. The sizes of the input feature maps are $[640\times192, 320\times96, 160\times48, 80\times24]$ and those of the patches are $[32, 16, 8, 4]$. Similarly, following the cascaded ITBs, four up-sampling (UP) layers are employed to recover the image resolution with strides 2, 4, 8, 16, generating feature maps with the same size as the original image.

As for \textbf{OMDA}, similar to~\cite{yan2018second}, we first generate a set of object-level point-clouds by cropping the points from the ground truth bounding boxes in the training data. Thereafter, we randomly select a subset of object-level point-clouds and paste them to a given LiDAR frame. With regard to contrastive learning, the temperature parameter $\tau$ is empirically fixed as 0.07. As to the memory bank, the momentum update hyper-parameter $m$ is set to 0.999.

In \textbf{3D Box Generation and Point Segmentation as well as Training Losses}, as displayed in \cref{fig:2} from the main paper, we follow the existing work~\cite{shi2019pointrcnn} and  introduce point segmentation as an auxiliary task by employing an extra segmentation head ${H}_{seg}(\cdot)$, which is trained by the segmentation loss $\mathcal{L}_{seg}$. Due to space limit, we omit the description on ${H}_{seg}(\cdot)$ and $\mathcal{L}_{seg}$ in the main paper for succinctness. In this document, we provide more details. 

Specifically, ${H}_{seg}(\cdot)$ consists of two fully connected (FC) layers, which is trained by the segmentation loss formulated as below:
\begin{equation}\label{segloss}
\mathcal{L}_{seg}=\sum_{\bm{p}_{i}}\mathcal{L}_{focal}(\bm{p}_{i}),\\
\end{equation}
where
\begin{equation}\label{focalloss}
\mathcal{L}_{focal}(\bm{p}_{i}) = -\alpha(1 - p')^\gamma \log(p')\\
\end{equation}
is the focal loss~\cite{lin2017focal}, and $\bm{p}_{i}$ is the $i$-th point. $p'$ equals to $p$ if $\bm{p}_{i}$ is the foreground point, and equals to $1-p$ otherwise, where $p$ is the predicted confidence score. During training, we keep the default setting, \emph{i.e.} $\alpha=0.25$ and $\gamma=2$ in Eq.~\eqref{focalloss}. Note that the ground-truth segmentation mask is naturally provided by the labels, \emph{i.e.} 3D points inside ground truth 3D boxes are considered as foreground points.

With point segmentation, a box regression head is introduced to generate 3D bounding box proposals. The feature for each proposal is obtained by randomly selecting 512 points in the corresponding proposal on top of the last layer of our two-stream multi-modal transformer. Subsequently, the refinement network consisting of three set abstraction (SA) layers is adopted to build a global representation, following which two cascaded $1\times1$ convolution layers for classification and regression are used to generate the prediction for detection including a 3D bounding box $(x, y, z, h, w, l, \theta)$ and a class confidence score $c$. Here, $(x, y, z)$ indicates the 3D coordinate of the object center, $(h, w, l)$ refers to the bounding box size, and $\theta$ is the orientation from the bird’s eye view. $\mathcal{L}_{rpn}$ includes the point-cloud segmentation loss $\mathcal{L}_{seg}$ and the proposal generation loss $\mathcal{L}_{pg}$, \emph{i.e.} $\mathcal{L}_{rpn}=\mathcal{L}_{seg}+\mathcal{L}_{pg}$. $\mathcal{L}_{pg}$ and $\mathcal{L}_{rcnn}$ denote the training objectives for the 3D proposal generation and refinement network, both of which consist of a classification loss and a regression loss. Concretely, for $(z, h, w, l)$, we directly utilize the smooth $L1$ loss for regression. For $(x, y, \theta)$, we use the bin-based loss~\cite{shi2019pointrcnn}. The overall 3D bounding box regression loss for the $i-$th bounding box is formulated as below:
\begin{align}\centering
\label{all_rcnn}
\mathcal{L}_\text{bin}^{(i)} = &\sum_{u \in \{(x, y, \theta)\}} ( \mathcal{L}_\text{ce}(\widehat{\text{bin}}_u^{(i)}, \text{bin}_u^{(i)}) \nonumber
\\
&+ \mathcal{L}_{\text{smooth-$L1$}}(\widehat{\text{res}}_u^{(i)}, \text{res}_u^{(i)}) ), \nonumber \\
\mathcal{L}_\text{res}^{(i)} = &\sum_{v \in \{(z, h, w, l)\}} \mathcal{L}_{\text{smooth-$L1$}}(\widehat{\text{res}}_v^{(i)}, \text{res}^{(i)}_v),  \\
\mathcal{L}_{{\text{box}}}^{(i)} = &\mathcal{L}_\text{bin}^{(i)} + \mathcal{L}_\text{res}^{(i)}, \nonumber
\end{align}
\noindent
where $\widehat {{\text{bin}}}^{(i)}$ and $\widehat {{\text{res}}}^{(i)}$ are the predicted bin assignments and residuals, respectively.  ${\text{bin}}^{(i)}$ and ${\text{res}}^{(i)}$ are the ground-truth targets, ${\mathcal{L}_\text{ce}}$ and $\mathcal{L}_{\text{smooth-$L1$}}$ denote the cross-entropy classiﬁcation loss and the smooth-$L1$ loss, respectively. Based on Eq.~(3), $\mathcal{L}_{rcnn}$ is written as the following:
\begin{equation}
{\mathcal{L}_{rcnn}} = \frac{1}{{\left\| {\rm B} \right\|}}\sum\limits_{i \in {\rm B}} {{\mathcal{L}_{ce}}(pro{b_i},labe{l_i})} {\rm{ + }}\frac{1}{{\left\| {{{\rm B}_{pos}}} \right\|}}\sum\limits_{i \in {{\rm B}_{pos}}} {\mathcal{L}_{{\rm{box}}}^{(i)}}
\end{equation}where $\rm B$ is the set of 3D proposals from RPN and $\rm B_{pos}$ is the set of positive proposals for regression. $prob_i$ is the conﬁdence score and $label_i$ refers to the corresponding ground-truth label. $\mathcal{L}_{pg}$ has the similar formulation as $\mathcal{L}_{rcnn}$.

\section{More Ablation Results}
\label{sec:B}

In this section, we provide more ablation results w.r.t. the CMT module, the memory bank size and the trade-off-parameter $\lambda$. 

As for \textbf{CMT}, we adopt this module after each PTB and ITB in four distinct levels as in \cref{fig:2} of the main paper, denoted by Layer-1, Layer-2, Layer-3, and Layer-4, respectively. We also adopt it between FPs in Pointformer and UPs in Imageformer, denoted as Layer-5. To validate the benefit of fully using CMT in all layers, we perform the ablation study by separately removing CMT from each layer. As summarized in Table \ref{table:A}, removing CMT at an arbitrary level deteriorates the performance, and CMT plays a more important role in higher levels based on the observation that the performance decreases more sharply when they are removed in Layer-3$\sim$5 than that in Layer-1$\sim$2. The results in Table \ref{table:A} also suggest that CMT can integrate multi-modal information in different levels, thus reaching the best performance when being fully used.

\begin{table}[t]
\centering
\resizebox{0.99\columnwidth}{!}{
\begin{tabular}{c|ccc|c}
\hline
\multicolumn{1}{c|}{Method} &\multicolumn{4}{|c}{3D Object Detection (\%)}\\
\hline
Levels&Car	&Ped.	&Cyc.	&mAP	\\
\hline
Fully used in Layer-1$\sim$5  &\textbf{83.58}	&\textbf{66.45}	&\textbf{76.22}	&\textbf{75.42}\\
Removed in Layer-1 &83.35	&66.18	&75.93	&75.15\\
Removed in Layer-2 &83.21	&65.83	&75.54	&74.86\\
Removed in Layer-3 &82.84	&65.30	&75.12	&74.42\\
Removed in Layer-4 &83.26	&65.14	&75.25	&74.55\\
Removed in Layer-5 &82.73	&64.89	&74.76	&74.13\\
\hline
\end{tabular}
}
\caption{Ablation results on the effect of CMT in different layers. }
\label{table:A}
\end{table}

We further explore the effect of the \textbf{memory bank size} on the performance of OMDA, by varying it from 256 to 4,096. It is worth noting that the bank size determines the number of negative samples used. As shown in Table \ref{table:B}, the mAP of CAT-Det increases as the bank size becomes larger, and reach the highest one when the size is 1,024. The reason behind lies in that the performance of OMDA increases when properly using more negative samples, but will be deteriorated when using too much negative pairs, since it probably leads to severe imbalance of positive/negative samples. We empirically set it to 1,024 in all the experiments.

\begin{figure}[t]
  \centering
   \includegraphics[width=0.99\linewidth]{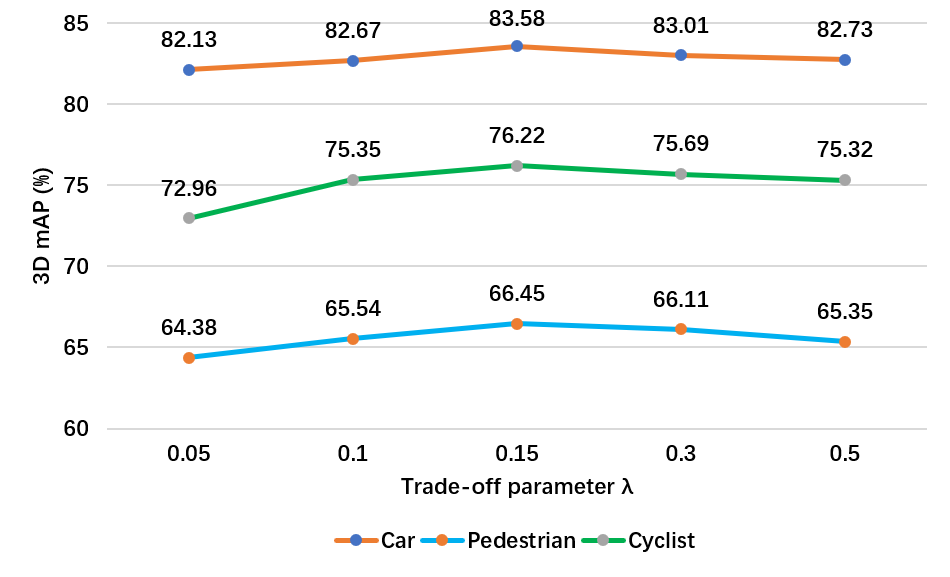}
   \caption{Ablation results on the trade-off parameter $\lambda$, which balances the effects of the detection loss  $(\mathcal{L}_{rpn}+\mathcal{L}_{rcnn})$ and the contrastive learning loss $(\mathcal{L}_{cl-p}+\mathcal{L}_{cl-o})$.}
   \label{fig:B}
\end{figure}

In regard of the \textbf{hyper-parameter} \textbf{$\bm{\lambda}$}, as shown in \cref{fig:B}, the contrastive learning loss is not fully used for supervision when $\lambda$ is small, thus yielding worse performance. In contrast, when $\lambda$ becomes too large, the credit of the detection loss is improperly suppressed, also incurring poor results. The two kinds of losses, \emph{i.e.} $(\mathcal{L}_{rpn}+\mathcal{L}_{rcnn})$ and $(\mathcal{L}_{cl-p}+\mathcal{L}_{cl-o})$, achieve their optimal balance when $\lambda=0.15$, which is therefore used as a default in our work.

\begin{table}[t]
\centering
\resizebox{0.85\columnwidth}{!}{
\begin{tabular}{c|ccc|c}
\hline
\multicolumn{1}{c|}{Memory Bank} &\multicolumn{4}{|c}{3D Object Detection (\%)}\\
\hline
Sizes&Car	&Ped.	&Cyc.	&mAP	\\
\hline
256&83.25	&66.15	&76.03	&75.14	\\
512&83.32	&66.14	&76.12	&75.19\\
1024&\textbf{83.58}	&\textbf{66.45}	&76.22	&\textbf{75.42}\\
2048&83.26	&66.21	&\textbf{76.37}	&75.28\\
4096&82.97	&66.13	&76.09	&75.06\\
\hline
\end{tabular}
}
\caption{Ablation results by using various memory bank sizes in the OMDA module.}
\label{table:B}
\end{table}

\begin{figure*}[!t]
\centering
\includegraphics[width=0.95\textwidth]{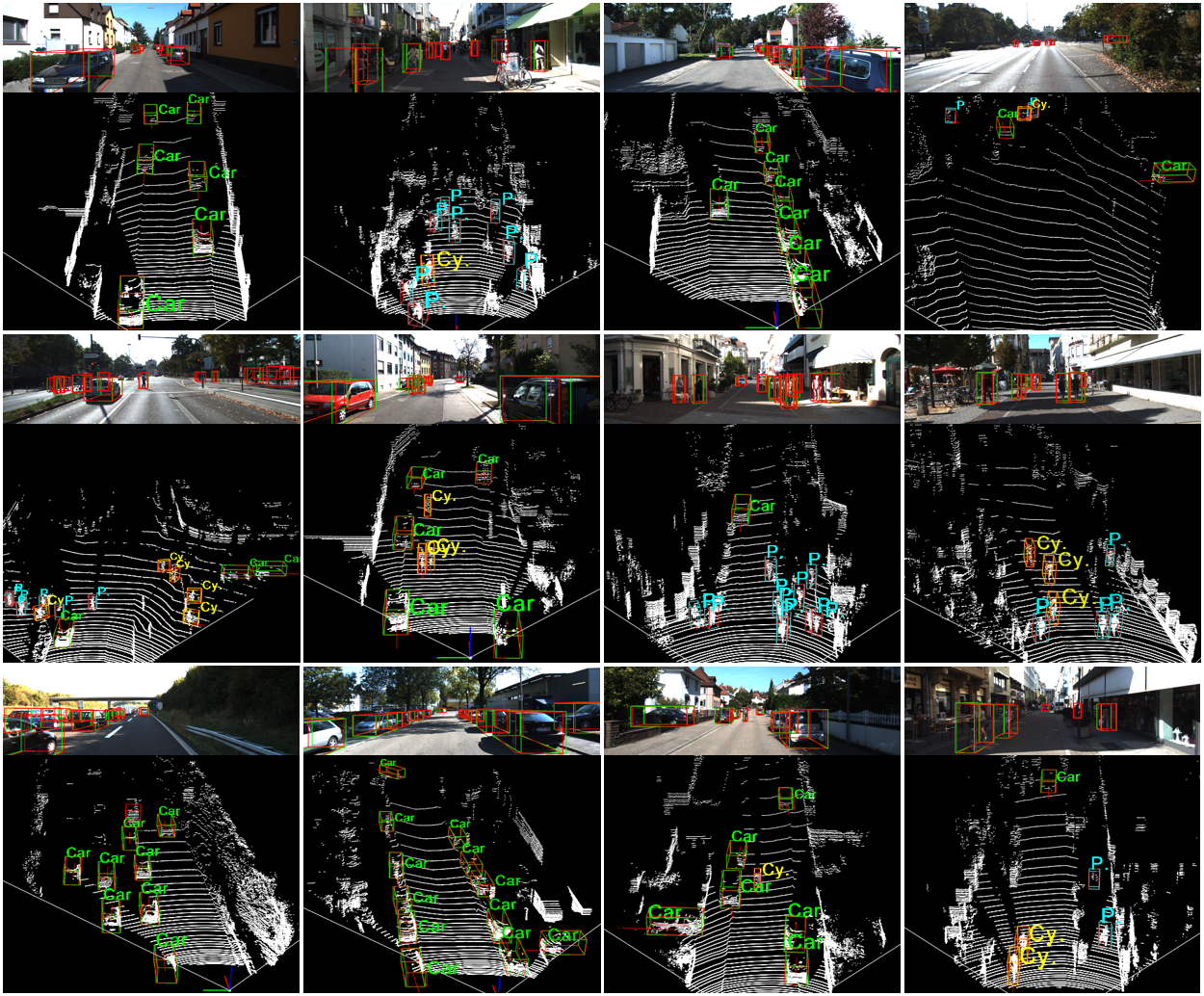} 
\caption{Visualized results by CAT-Det on KITTI val split. Red/green rectangles indicate predicted/GT bounding boxes.}
\label{fig:C}
\end{figure*}

We also add results of applying \textbf{GT-Paste} only on point-clouds in Table \ref{table:C} and show that inconsistent data augmentation tends to degrade the results (even worse than that without GT-Paste). Instead, OMDA well addresses this issue.

\begin{table}[t]
\centering
\resizebox{0.95\columnwidth}{!}{
\begin{tabular}{c|ccc|c}
\hline
Method &Car	&Ped.	&Cyc.	&mAP(\%)	\\
\hline
w/o GT-Paste &81.73	&63.30	&70.63	&71.89\\
with GT-Paste on 3D points &81.13	&61.15	&69.42	&70.57\\
with OMDA &83.58	&66.45	&76.22	&75.42\\
\hline
\end{tabular}
}
\caption{Ablation results on GT-Paste and contrastive learning on KITTI Val. }
\label{table:C}
\end{table}

In order to further investigate the contributions of ITB we add more results by \textbf{removing ITB and local/global transformers}. As PTB cannot be directly removed, we replace it by PointNet++. The results summarized in Table \ref{table:D} highlight their effectiveness.

\begin{table}[t]
\centering
\resizebox{0.95\columnwidth}{!}{
\begin{tabular}{c|ccc|c}
\hline
Method &Car	&Ped.	&Cyc.	&mAP(\%)	\\
\hline
Full model &83.58	&66.45	&76.22	&75.42\\
w/o ITB &81.22	&63.20	&70.18	&71.53\\
Replacing PTB by PointNet++ &82.69	&64.93	&74.61	&74.08\\
w/o Global Transformer &82.83	&65.21	&74.88	&74.31\\
w/o Local Transformer &83.19	&65.94	&75.65	&74.93\\
\hline
\end{tabular}
}
\caption{Ablation results on ITB and PTB on KITTI Val. }
\label{table:D}
\end{table}

Finally, we report the \textbf{runtime and the size of parameters} with comparisons to major multi-modal methods and transformer based ones. As in Table \ref{table:E}, our method reaches a good trade-off, with the highest accuracy and moderately increased model size and inference time. In the future, we will explore compression techniques to reduce the complexity of the transformer based methods for practical use.

\begin{table}[t]
\centering
\resizebox{0.95\columnwidth}{!}{
\begin{tabular}{c | c | c | c | c}
\hline
{Method} & {Modality} & {Params~(M)} & {Time~(ms)} & {mAP~(\%)} \\ \hline
AVOD-FPN  & L+I & 38.07 & 100 & 56.84\\
F-PointNet  & L+I & 12.45 & 167 & 57.86\\
EPNet  & L+I & 16.23 & 178 & -- \\
VPFNet~\cite{wang2021vpfnet}  & L+I & -- & 200 & 65.99 \\
\hline
PointTransformer & L & 6.06  & 250  & 61.66\\
M3DETR  & L & 19.66 & 256 & 64.65\\
\hline
CAT-Det (\bf Ours)  & L+I & 23.21  & 314  & 67.05 \\
\hline
\end{tabular}
}
\caption{Comparison of mAP, size of parameters and runtime on KITTI Test.}
\label{table:E}
\end{table}

\begin{table}[!t]
\centering
\resizebox{0.95\columnwidth}{!}{
\begin{tabular}{c | c | c | c}
\hline
{\bf Setting} & {\bf Easy} & {\bf Moderate} & {\bf Hard}\\ \hline
Car (Detection) & 95.97 & 94.71 & 92.07\\
Car (Orientation) & 95.95 & 94.57 & 91.88\\
Car (3D Detection) & 89.87 & 81.32 & 76.68\\
Car (Bird's Eye View) & 92.59 & 90.07 & 85.82\\
Pedestrian (Detection) & 67.15 & 56.75 & 53.44\\
Pedestrian (Orientation) & 52.75 & 43.86 & 41.15\\
Pedestrian (3D Detection) & 54.26 & 45.44 & 41.94\\
Pedestrian (Bird's Eye View) & 57.13 & 48.78 & 45.56\\
Cyclist (Detection) & 87.94 & 80.70 & 73.86\\
Cyclist (Orientation) & 87.79 & 80.25 & 73.41\\
Cyclist (3D Detection) & 83.68 & 68.81 & 61.45\\
Cyclist (Bird's Eye View) & 85.35 & 72.51 & 65.55\\
\hline
\end{tabular}
}
\caption{Detailed results (\%) on the official KITTI test leaderboard by using CAT-Det.}
\label{table:F}
\end{table}

\section{More Visualization Results}
\label{sec:C}
As in \cref{fig:7} in the main paper, we demonstrate a few visualization results by performing 3D detection on KITTI val split via CAT-Det. In this section, we display more results.

As shown in \cref{fig:C}, our approach precisely predicts both locations and orientations of 3D objects even under extremely challenging situations, including remote objects (the top row), tiny objects (the middle row) and objects with heavy occlusions (the bottom row).

\section{Details on Official KITTI Test Leaderboard}
\label{sec:D}
\begin{figure*}[!t]
    \centering
    \begin{tabular}{c|c|c}
    \Large Car & \Large Pedestrian & \Large Cyclist \\ \midrule
    \begin{overpic}[width=0.30\linewidth,trim={1.8cm 0.9cm 1.4cm 1.2cm},clip]
    {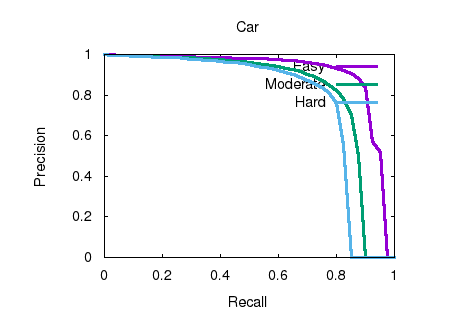}
        \put (20,43){\textbf{3D@0.7}}
        \put (20,36){\textbf{CAT-Det}}
        \put (20,29){\color{Purple}   \textbf{Easy: 89.87}}
        \put (20,22){\color{Green}    \textbf{Mode:81.32}} 
        \put (20,15){\color{RoyalBlue}\textbf{Hard: 76.68}}
    \end{overpic} &
    \begin{overpic}[width=0.30\linewidth,trim={1.8cm 0.9cm 1.4cm 1.2cm},clip]
    {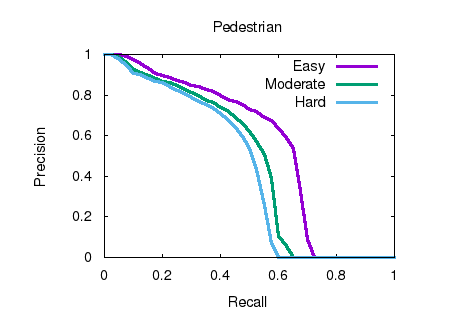}
        \put (20,43){\textbf{3D@0.5}}
        \put (20,36){\textbf{CAT-Det}}
        \put (20,29){\color{Purple}   \textbf{Easy: 54.26}}
        \put (20,22){\color{Green}    \textbf{Mode:45.44}}
        \put (20,15){\color{RoyalBlue}\textbf{Hard: 41.94}}
    \end{overpic} &
    \begin{overpic}[width=0.30\linewidth,trim={1.8cm 0.9cm 1.4cm 1.2cm},clip]
    {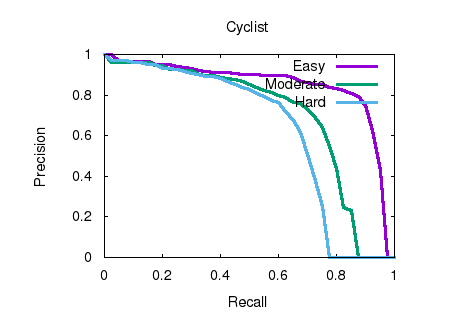}
        \put (20,43){\textbf{3D@0.5}}
        \put (20,36){\textbf{CAT-Det}}
        \put (20,29){\color{Purple}   \textbf{Easy: 83.68}}
        \put (20,22){\color{Green}    \textbf{Mode:68.81}}
        \put (20,15){\color{RoyalBlue}\textbf{Hard: 61.45}}
    \end{overpic} \\ \midrule

    \begin{overpic}[width=0.30\linewidth,trim={1.8cm 0.9cm 1.4cm 1.2cm},clip]
    {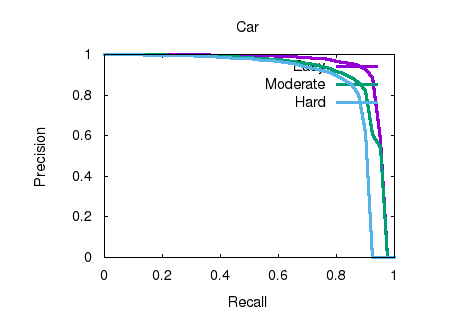}
        \put (20,43){\textbf{BEV@0.7}}
        \put (20,36){\textbf{CAT-Det}}
        \put (20,29){\color{Purple}   \textbf{Easy: 92.59}}
        \put (20,22){\color{Green}    \textbf{Mode:90.07}} 
        \put (20,15){\color{RoyalBlue}\textbf{Hard: 85.82}}
    \end{overpic} &
    \begin{overpic}[width=0.30\linewidth,trim={1.8cm 0.9cm 1.4cm 1.2cm},clip]
    {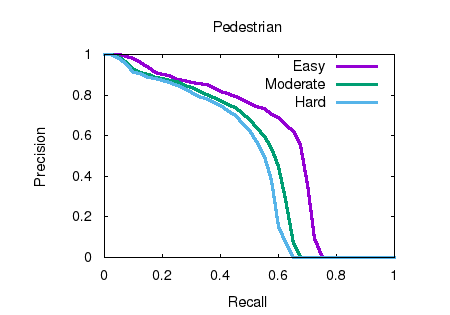}
        \put (20,43){\textbf{BEV@0.5}}
        \put (20,36){\textbf{CAT-Det}}
        \put (20,29){\color{Purple}   \textbf{Easy: 57.13}}
        \put (20,22){\color{Green}    \textbf{Mode:48.78}}
        \put (20,15){\color{RoyalBlue}\textbf{Hard: 45.56}}
    \end{overpic} &
    \begin{overpic}[width=0.30\linewidth,trim={1.8cm 0.9cm 1.4cm 1.2cm},clip]
    {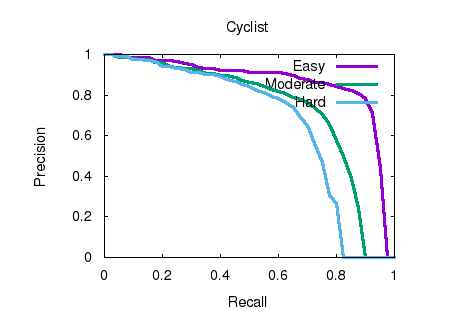}
        \put (20,43){\textbf{BEV@0.5}}
        \put (20,36){\textbf{CAT-Det}}
        \put (20,29){\color{Purple}   \textbf{Easy: 85.35}}
        \put (20,22){\color{Green}    \textbf{Mode:72.51}}
        \put (20,15){\color{RoyalBlue}\textbf{Hard: 65.55}}
    \end{overpic} \\ \midrule

    \begin{overpic}[width=0.30\linewidth,trim={1.8cm 0.9cm 1.4cm 1.2cm},clip]
    {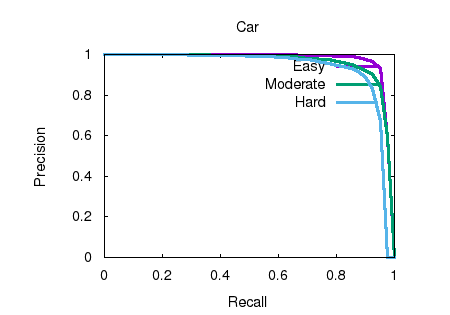}
        \put (20,43){\textbf{2D@0.7}}
        \put (20,36){\textbf{CAT-Det}}
        \put (20,29){\color{Purple}   \textbf{Easy: 95.97}}
        \put (20,22){\color{Green}    \textbf{Mode:94.71}}
        \put (20,15){\color{RoyalBlue}\textbf{Hard: 92.07}}
    \end{overpic} &
    \begin{overpic}[width=0.30\linewidth,trim={1.8cm 0.9cm 1.4cm 1.2cm},clip]
    {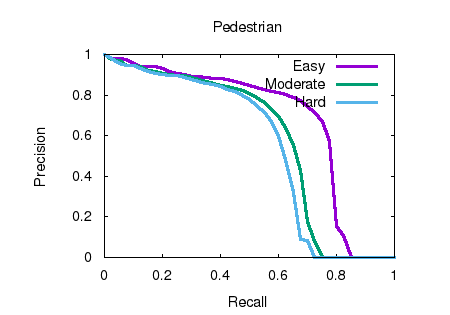}
        \put (20,43){\textbf{2D@0.5}}
        \put (20,36){\textbf{CAT-Det}}
        \put (20,29){\color{Purple}   \textbf{Easy: 67.15}} 
        \put (20,22){\color{Green}    \textbf{Mode:56.75}}
        \put (20,15){\color{RoyalBlue}\textbf{Hard: 53.44}}
    \end{overpic} &
    \begin{overpic}[width=0.30\linewidth,trim={1.8cm 0.9cm 1.4cm 1.2cm},clip]
    {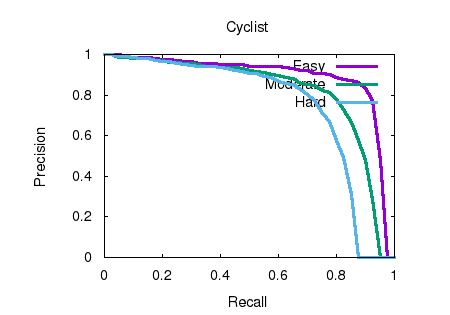}
        \put (20,43){\textbf{2D@0.5}}
        \put (20,36){\textbf{CAT-Det}}
        \put (20,29){\color{Purple}   \textbf{Easy: 87.94}}
        \put (20,22){\color{Green}    \textbf{Mode:80.70}}
        \put (20,15){\color{RoyalBlue}\textbf{Hard: 73.86}}
    \end{overpic} \\ \midrule

    \begin{overpic}[width=0.30\linewidth,trim={1.8cm 0.9cm 1.4cm 1.2cm},clip]
    {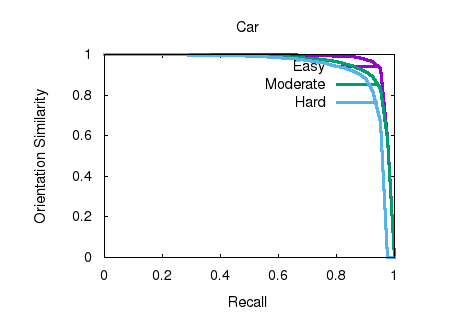}
        \put (20,43){\textbf{AOS}}
        \put (20,36){\textbf{CAT-Det}}
        \put (20,29){\color{Purple}   \textbf{Easy: 95.95}}
        \put (20,22){\color{Green}    \textbf{Mode:94.57}}
        \put (20,15){\color{RoyalBlue}\textbf{Hard: 91.88}}
    \end{overpic} &
    \begin{overpic}[width=0.30\linewidth,trim={1.8cm 0.9cm 1.4cm 1.2cm},clip]
    {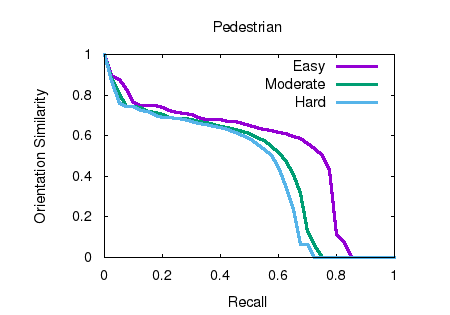}
        \put (20,43){\textbf{AOS}}
        \put (20,36){\textbf{CAT-Det}}
        \put (20,29){\color{Purple}   \textbf{Easy: 52.75}}
        \put (20,22){\color{Green}    \textbf{Mode:43.86}}
        \put (20,15){\color{RoyalBlue}\textbf{Hard: 41.15}}
    \end{overpic} &
    \begin{overpic}[width=0.30\linewidth,trim={1.8cm 0.9cm 1.4cm 1.2cm},clip]
    {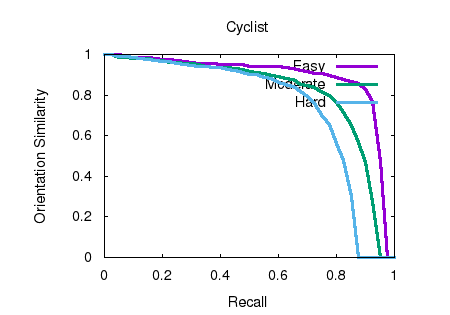}
        \put (20,43){\textbf{AOS}}
        \put (20,36){\textbf{CAT-Det}}
        \put (20,29){\color{Purple}   \textbf{Easy: 87.79}}
        \put (20,22){\color{Green}    \textbf{Mode:80.25}}
        \put (20,15){\color{RoyalBlue}\textbf{Hard: 73.41}}
    \end{overpic}
	\end{tabular}
    \caption{
    \textbf{Precision Recall Curves} on the official KITTI test leaderboard by using CAT-Det.
    \textbf{Left to right:} Car@0.7, Pedestrian@0.5 and Cyclist@0.5.
    \textbf{Top to bottom:} 3D Detection, Bird's Eye View, 2D Detection and Orientation.}
	\label{fig:D}
\end{figure*}

In Table \ref{table:1} and Table \ref{table:2} from the main paper, we summarize the state-of-the-art results w.r.t AP/mAP on KITTI val/test splits. In this section, we provide more detailed results. Table~\ref{table:F} shows the official results in various settings (\emph{i.e.} 3D, BEV, 2D and AOS) for three distinct levels of difficulties from the KITTI leaderboard. We also present the Precision-Recall curves in \cref{fig:D} on the test set.

\end{document}